\pdfoutput=1

\documentclass{article}

\usepackage[final, nonatbib]{nips_2016}
\usepackage{color}

\usepackage[utf8]{inputenc} 
\usepackage[T1]{fontenc}    
\usepackage{hyperref}       
\usepackage{url}            
\usepackage{booktabs}       
\usepackage{amsfonts}       
\usepackage{nicefrac}       
\usepackage{microtype}      

\usepackage{algorithm, algpseudocode}
\usepackage{mathrsfs}
\usepackage{dsfont}
\usepackage{lmodern}
\usepackage{array}
\usepackage{amsmath}
\usepackage{amssymb}
\usepackage{graphicx}
\usepackage{mathrsfs}
\usepackage{psfrag}
\usepackage{color}
\usepackage{here}
\usepackage{wasysym}
\usepackage{enumitem}
\usepackage{xcolor}
\usepackage{caption}
\usepackage{subcaption}

\newcommand{\Exp}{\mathds{E}}

\newcommand{\Nat}{\mathbb{N}}
\newcommand{\Ind}{\mathds{1}}

\begin{document}

\title{Deep Exploration via Bootstrapped DQN}

\author{
Ian Osband$^{1,2}$, Charles Blundell$^{2}$, Alexander Pritzel$^{2}$, Benjamin Van Roy$^{1}$\\
$^1$Stanford University, $^2$Google DeepMind \\
\texttt{\{iosband, cblundell, apritzel\}@google.com, bvr@stanford.edu}
}

\maketitle


\vspace{-4mm}

\begin{abstract}

Efficient exploration remains a major challenge for reinforcement learning (RL).
Common dithering strategies for exploration, such as $\epsilon$-greedy, do not carry out temporally-extended (or deep) exploration; this can lead to exponentially larger data requirements.
However, most algorithms for statistically efficient RL are not computationally tractable in complex environments.
Randomized value functions offer a promising approach to efficient exploration with generalization, but existing algorithms are not compatible with nonlinearly parameterized value functions.
As a first step towards addressing such contexts we develop \textit{bootstrapped DQN}.
We demonstrate that bootstrapped DQN can combine deep exploration with deep neural networks for exponentially faster learning than any dithering strategy.
In the Arcade Learning Environment bootstrapped DQN substantially improves learning speed and cumulative performance across most games.

\end{abstract}

\section{Introduction}
\label{sec: intro}

We study the reinforcement learning (RL) problem where an agent interacts with an unknown environment.
The agent takes a sequence of actions in order to maximize cumulative rewards.
Unlike standard planning problems, an RL agent does not begin with perfect knowledge of the environment, but learns through experience.
This leads to a fundamental trade-off of exploration versus exploitation;
the agent may improve its future rewards by exploring poorly understood states and actions, but this may require sacrificing immediate rewards.
To learn efficiently an agent should explore only when there are valuable learning opportunities.
Further, since any action may have long term consequences, the agent should reason about the informational value of possible observation sequences.
Without this sort of temporally extended (deep) exploration, learning times can worsen by an exponential factor.

The theoretical RL literature offers a variety of provably-efficient approaches to deep exploration \cite{Jaksch2010}.
However, most of these are designed for Markov decision processes (MDPs) with small finite state spaces, while others require solving computationally intractable planning tasks \cite{guez2012efficient}.
These algorithms are not practical in complex environments where an agent must generalize to operate effectively.
For this reason, large-scale applications of RL have relied upon statistically inefficient strategies for exploration \cite{mnih2015human} or even no exploration at all \cite{tesauro1995temporal}.
We review related literature in more detail in Section \ref{sec: related}.

Common dithering strategies, such as $\epsilon$-greedy, approximate the value of an action by a single number.
Most of the time they pick the action with the highest estimate, but sometimes they choose another action at random.
In this paper, we consider an alternative approach to efficient exploration inspired by Thompson sampling.
These algorithms have some notion of uncertainty and instead maintain a \textit{distribution} over possible values.
They explore by randomly select a policy according to the probability it is the optimal policy.
Recent work has shown that randomized value functions can implement something similar to Thompson sampling without the need for an intractable exact posterior update.
However, this work is restricted to linearly-parameterized value functions \cite{osband2014generalization}.
We present a natural extension of this approach that enables use of complex non-linear generalization methods such as deep neural networks.
We show that the bootstrap with random initialization can produce reasonable uncertainty estimates for neural networks at low computational cost.
Bootstrapped DQN leverages these uncertainty estimates for efficient (and deep) exploration.
We demonstrate that these benefits can extend to large scale problems that are not designed to highlight deep exploration.
Bootstrapped DQN substantially reduces learning times and improves performance across most games.
This algorithm is computationally efficient and parallelizable; on a single machine our implementation runs roughly $20$\% slower than DQN.

\section{Uncertainty for neural networks}
\label{sec: uncertainty nn}

Deep neural networks (DNN) represent the state of the art in many supervised and reinforcement learning domains \cite{mnih2015human}.
We want an exploration strategy that is statistically computationally efficient together with a DNN representation of the value function.
To explore efficiently, the first step to quantify uncertainty in value estimates so that the agent can judge potential benefits of exploratory actions.
The neural network literature presents a sizable body of work on uncertainty quantification founded on parametric Bayesian inference \cite{blundell2015weight,gal2015dropout}.
We actually found the simple non-parametric bootstrap with random initialization \cite{efron1982jackknife} more effective in our experiments, but the main ideas of this paper would apply with any other approach to uncertainty in DNNs.

The bootstrap princple is to approximate a population distribution by a sample distribution \cite{efron1994introduction}.
In its most common form, the bootstrap takes as input a data set $D$ and an estimator $\psi$.
To generate a sample from the bootstrapped distribution, a data set $\tilde{D}$ of cardinality equal to that of $D$ is sampled uniformly with replacement from $D$.
The bootstrap sample estimate is then taken to be $\psi(\tilde{D})$.
The bootstrap is widely hailed as a great advance of 20th century applied statistics and even comes with theoretical guarantees \cite{bickel1981some}.
In Figure \ref{fig: shared convnet} we present an efficient and scalable method for generating bootstrap samples from a large and deep neural network.
The network consists of a shared architecture with $K$ bootstrapped ``heads'' branching off independently.
Each head is trained only on its bootstrapped sub-sample of the data and represents a single bootstrap sample $\psi(\tilde{D})$.
The shared network learns a joint feature representation across all the data, which can provide significant computational advantages at the cost of lower diversity between heads.
This type of bootstrap can be trained efficiently in a single forward/backward pass; it can be thought of as a data-dependent dropout, where the dropout mask for each head is fixed for each data point \cite{srivastava2014dropout}.

{\small
\begin{figure}[!h]
   \centering
   \vspace{-2mm}
   \begin{subfigure}[b]{0.31\linewidth}
       \includegraphics[height=1.1in]{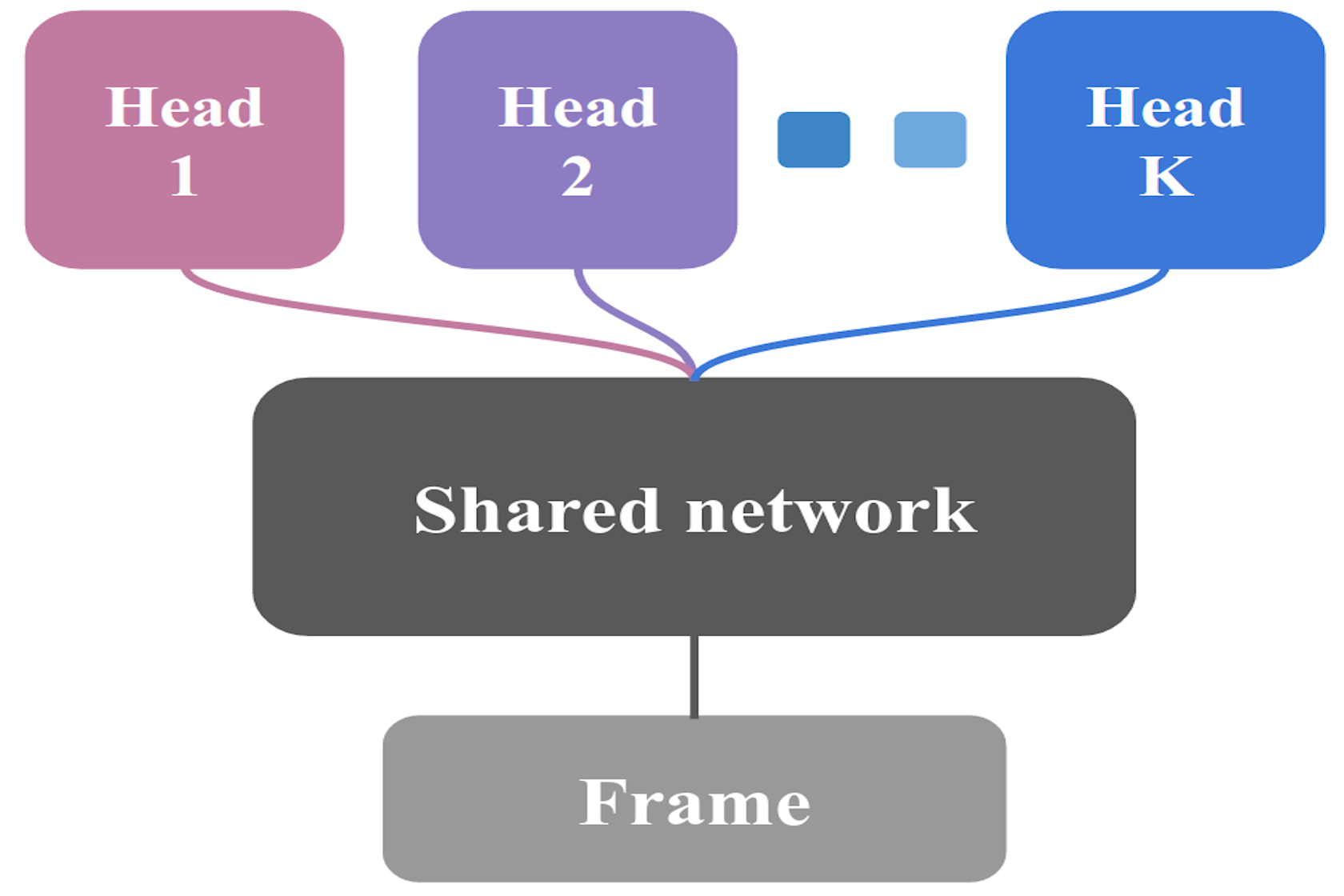}
       \caption{Shared network architecture}
       \label{fig: shared convnet}
   \end{subfigure}
   \hspace{1mm}
   \begin{subfigure}[b]{0.31\linewidth}
       \includegraphics[height=1.1in]{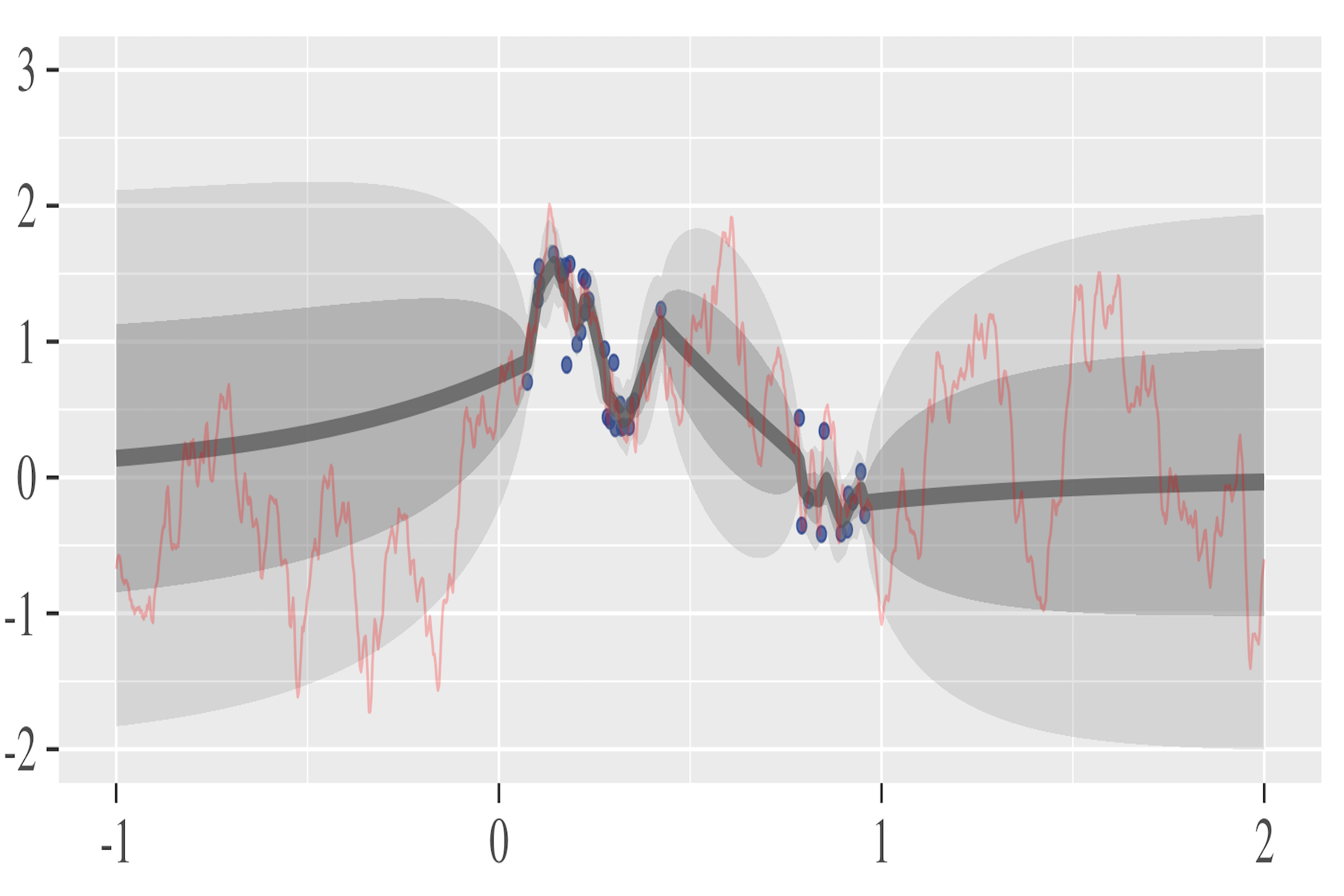}
       \caption{Gaussian process posterior}
       \label{fig: gp regression}
   \end{subfigure}
   \hspace{1mm}
   \begin{subfigure}[b]{0.31\linewidth}
       \includegraphics[height=1.1in]{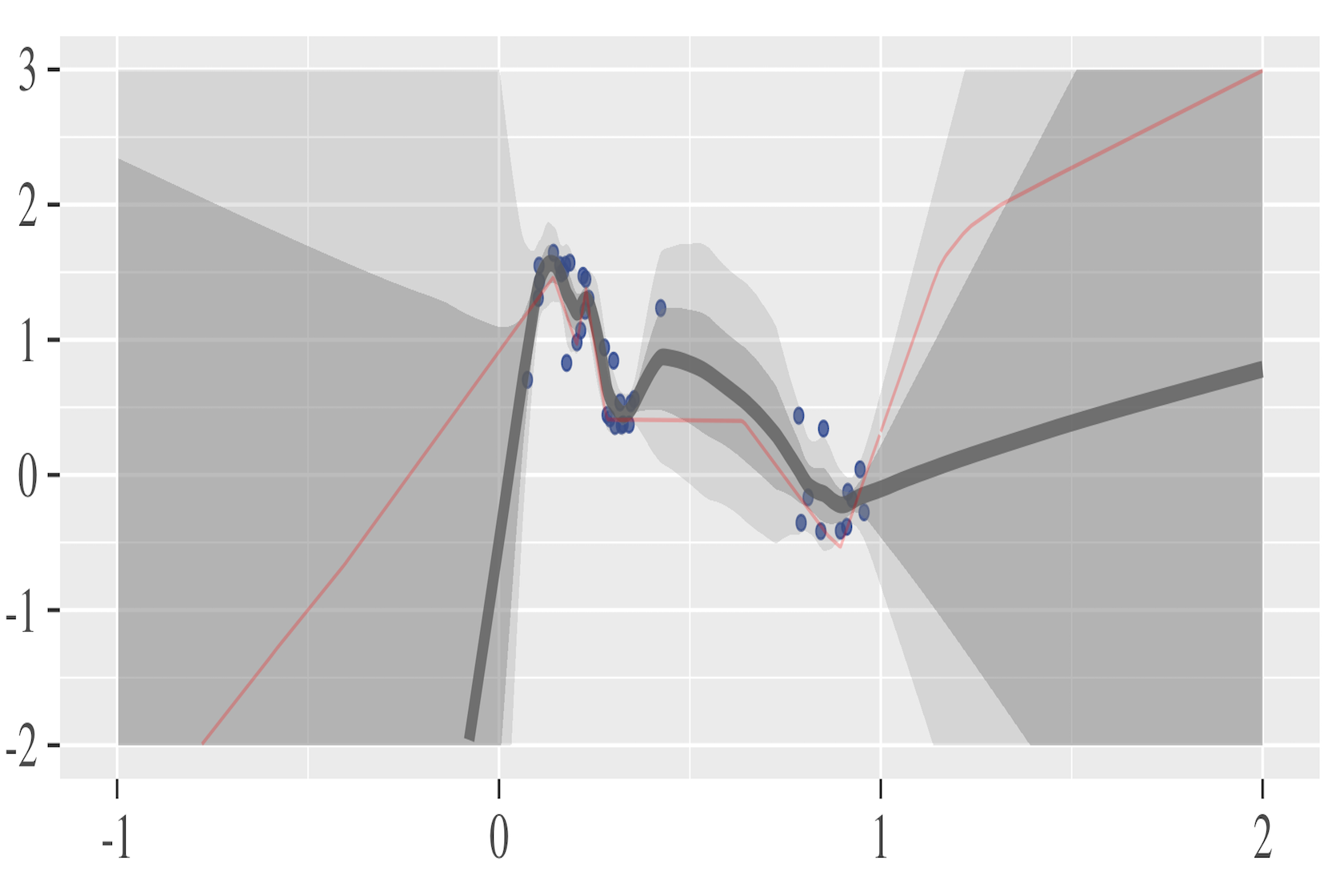}
       \caption{Bootstrapped neural nets}
       \label{fig: bootstrap regression}
   \end{subfigure}
   \caption{\small Bootstrapped neural nets can produce reasonable posterior estimates for regression.}
   \vspace{-2mm}
   \label{fig: regression posteriors}
\end{figure}
}

Figure \ref{fig: regression posteriors} presents an example of uncertainty estimates from bootstrapped neural networks on a regression task with noisy data.
We trained a fully-connected 2-layer neural networks with 50 rectified linear units (ReLU) in each layer on 50 bootstrapped samples from the data.
As is standard, we initialize these networks with random parameter values, this induces an important initial diversity in the models.
We were unable to generate effective uncertainty estimates for this problem using the dropout approach in prior literature \cite{gal2015dropout}.
Further details are provided in Appendix \ref{app: uncertainty nn}.

\vspace{-3mm}
\section{Bootstrapped DQN}
\label{sec: boostrapped DQN}
\vspace{-1mm}
For a policy $\pi$ we define the value of an action $a$ in state $s$
$Q^\pi(s,a) := \Exp_{s,a,\pi}\left[ \sum_{t=1}^\infty \gamma^t r_t \right]$,
where $\gamma \in (0,1)$ is a discount factor that balances immediate versus future rewards $r_t$.
This expectation indicates that the initial state is $s$, the initial action is $a$, and thereafter actions are selected by the policy $\pi$.
The optimal value is $Q^*(s,a) := \max_\pi Q^\pi(s,a)$.
To scale to large problems, we learn a parameterized estimate of the Q-value function $Q(s,a ; \theta)$ rather than a tabular encoding.
We use a neural network to estimate this value.

The Q-learning update from state $s_t$, action $a_t$, reward $r_t$ and new state $s_{t+1}$ is given by
\vspace{-1mm}
\begin{equation}
\label{eq:qlearn}
    \theta_{t+1} \leftarrow \theta_t + \alpha (y^Q_t - Q(s_t, a_t; \theta_t)) \nabla_\theta Q(s_t, a_t ; \theta_t)
\end{equation}
where $\alpha$ is the scalar learning rate and $y^Q_t$ is the target value $r_t + \gamma \max_a Q(s_{t+1}, a; \theta^-)$. $\theta^-$ are target network parameters fixed $\theta^-=\theta_t$.

Several important modifications to the Q-learning update improve stability for DQN \cite{mnih2015human}.
First the algorithm learns from sampled transitions from an experience buffer, rather than learning fully online.
Second the algorithm uses a target network with parameters $\theta^-$ that are
copied from the learning network $\theta^- \leftarrow \theta_t$ only every $\tau$ time steps and then kept fixed in between updates.
Double DQN \cite{van2015deep} modifies the target $y^Q_t$ and helps further\footnote{In this paper we use the DDQN update for all DQN variants unless explicitly stated.}:
\vspace{-1mm}
\begin{equation}
\label{eq:yqt}
  y^{Q}_t \leftarrow r_t + \gamma \max_a Q\big(s_{t+1}, \arg\max_a Q(s_{t+1}, a; \theta_t) ; \theta^-\big).
\end{equation}
Bootstrapped DQN modifies DQN to approximate a \textit{distribution} over Q-values via the bootstrap.
At the start of each episode, bootstrapped DQN samples a single Q-value function from its approximate posterior.
The agent then follows the policy which is optimal for that \textit{sample} for the duration of the episode.
This is a natural adaptation of the Thompson sampling heuristic to RL that allows for temporally extended (or deep) exploration \cite{Strens00,Osband2013}.

We implement this algorithm efficiently by building up $K \in \Nat$ bootstrapped estimates of the Q-value function in parallel as in Figure \ref{fig: shared convnet}.
Importantly, each one of these value function function heads $Q_k(s,a; \theta)$ is trained against its own target network $Q_k(s,a; \theta^-)$.
This means that each $Q_1, .., Q_K$ provide a temporally extended (and consistent) estimate of the value uncertainty via TD estimates.
In order to keep track of which data belongs to which bootstrap head we store flags $w_1, .., w_K \in \{0,1\}$ indicating which heads are privy to which data.
We approximate a bootstrap sample by selecting $k \in \{1, ..,K\}$ uniformly at random and following $Q_k$ for the duration of that episode.
We present a detailed algorithm for our implementation of bootstrapped DQN in Appendix \ref{app: boot dqn algorithm}.


\vspace{-1mm}
\section{Related work}
\label{sec: related}
\vspace{-2mm}
The observation that temporally extended exploration is necessary for efficient reinforcement learning is not new.
For any prior distribution over MDPs, the optimal exploration strategy is available through dynamic programming in the Bayesian belief state space.
However, the exact solution is intractable even for very simple systems\cite{guez2012efficient}.
Many successful RL applications focus on generalization and planning but address exploration only via inefficient exploration \cite{mnih2015human} or even none at all \cite{tesauro1995temporal}.
However, such exploration strategies can be highly inefficient.


Many exploration strategies are guided by the principle of ``optimism in the face of uncertainty'' (OFU).
These algorithms add an exploration bonus to values of state-action pairs that may lead to useful learning and select actions to maximize these adjusted values.
This approach was first proposed for finite-armed bandits \cite{lai1985asymptotically}, but the principle has been extended successfully across bandits with generalization and tabular RL \cite{Jaksch2010}.
Except for particular deterministic contexts \cite{WenVanroy13}, OFU methods that lead to efficient RL in complex domains have been computationally intractable.
The work of \cite{stadie2015incentivizing} aims to add an effective bonus through a variation of DQN.
The resulting algorithm relies on a large number of hand-tuned parameters and is only suitable for application to deterministic problems.
We compare our results on Atari to theirs in Appendix \ref{app: atari} and find that bootstrapped DQN offers a significant improvement over previous methods.


Perhaps the oldest heuristic for balancing exploration with exploitation is given by Thompson sampling \cite{Thompson1933}.
This bandit algorithm takes a single sample from the posterior at every time step and chooses the action which is optimal for that time step.
To apply the Thompson sampling principle to RL, an agent should sample a value function from its posterior.
Naive applications of Thompson sampling to RL which resample every timestep can be extremely inefficient.
The agent must also commit to this sample for several time steps in order to achieve deep exploration \cite{Strens00,guez2012efficient}.
The algorithm PSRL does exactly this, with state of the art guarantees \cite{Osband2013,osband2014model}.
However, this algorithm still requires solving a single known MDP, which will usually be intractable for large systems.


Our new algorithm, bootstrapped DQN, approximates this approach to exploration via randomized value functions sampled from an approximate posterior.
Recently, authors have proposed the RLSVI algorithm which accomplishes this for linearly parameterized value functions.
Surprisingly, RLSVI recovers state of the art guarantees in the setting with tabular basis functions, but its performance is crucially dependent upon a suitable linear representation of the value function \cite{osband2014generalization}.
We extend these ideas to produce an algorithm that can simultaneously perform generalization and exploration with a flexible nonlinear value function representation.
Our method is simple, general and compatible with almost all advances in deep RL at low computational cost and with few tuning parameters.

\section{Deep Exploration}
\label{sec: marshmallow}

Uncertainty estimates allow an agent to direct its exploration at potentially informative states and actions.
In bandits, this choice of directed exploration rather than dithering generally categorizes efficient algorithms.
The story in RL is not as simple, directed exploration is not enough to guarantee efficiency; the exploration must also be deep.
Deep exploration means exploration which is directed over multiple time steps; it can also be called ``planning to learn'' or ``far-sighted'' exploration.
Unlike bandit problems, which balance actions which are immediately rewarding or immediately informative, RL settings require planning over several time steps \cite{Kakade2003}.
For exploitation, this means that an efficient agent must consider the future rewards over several time steps and not simply the myopic rewards.
In exactly the same way, efficient exploration may require taking actions which are neither immediately rewarding, nor immediately informative.

To illustrate this distinction, consider a simple deterministic chain $\{s_{-3}, .., s_{+3}\}$ with three step horizon starting from state $s_0$.
This MDP is known to the agent a priori, with deterministic actions ``left'' and ``right''.
All states have zero reward, except for the leftmost state $s_{-3}$ which has known reward $\epsilon > 0$ and the rightmost state $s_{3}$ which is unknown.
In order to reach either a rewarding state or an informative state within three steps from $s_0$ the agent must plan a consistent strategy over several time steps.
Figure \ref{fig: deep explore} depicts the planning and look ahead trees for several algorithmic approaches in this example MDP.
The action ``left'' is gray, the action ``right'' is black.
Rewarding states are depicted as red, informative states as blue.
Dashed lines indicate that the agent can plan ahead for either rewards or information.
Unlike bandit algorithms, an RL agent can plan to exploit future rewards.
Only an RL agent with deep exploration can plan to learn.

{\footnotesize
\begin{figure}[!h]
   \centering
   \begin{subfigure}[b]{0.99\linewidth}
       \includegraphics[width=\linewidth]{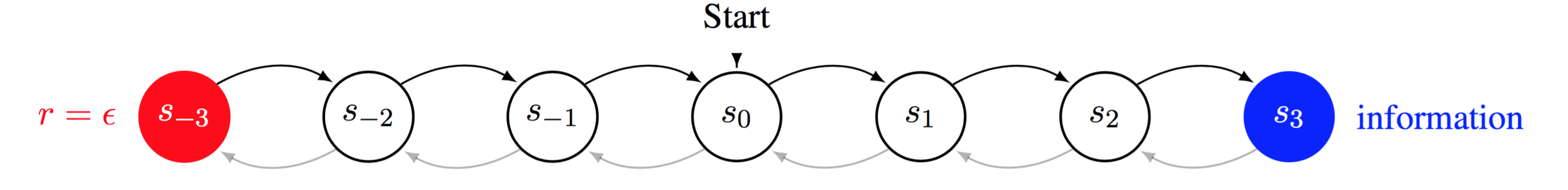}
   \end{subfigure}
   \begin{subfigure}[b]{0.23\linewidth}
       \includegraphics[width=\linewidth]{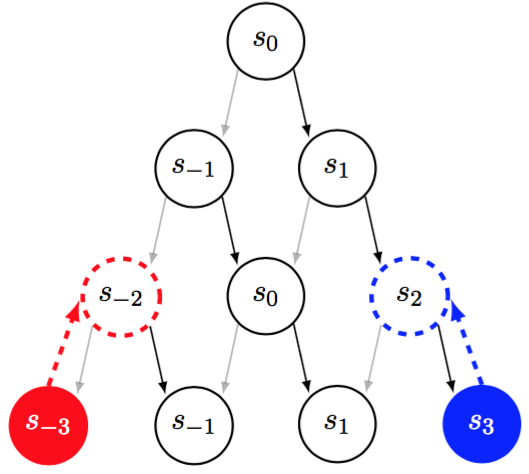}
       \caption{Bandit algorithm}
       \label{fig: rl bandit}
   \end{subfigure}
   \hspace{1mm}
   \begin{subfigure}[b]{0.23\linewidth}
       \includegraphics[width=\linewidth]{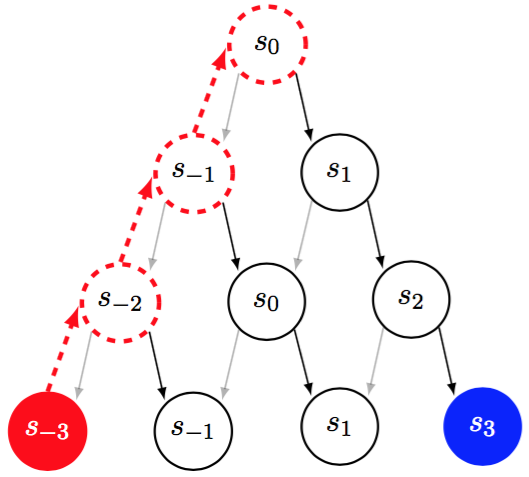}
       \caption{RL+dithering}
       \label{fig: rl dither}
   \end{subfigure}
   \hspace{1mm}
   \begin{subfigure}[b]{0.23\linewidth}
       \includegraphics[width=\linewidth]{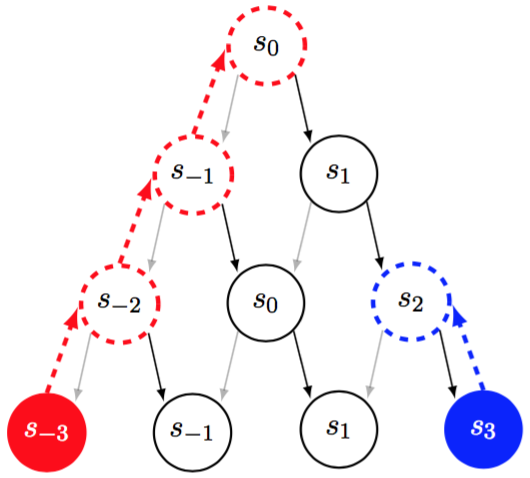}
       \caption{RL+shallow explore}
       \label{fig: rl shallow}
   \end{subfigure}
   \hspace{1mm}
   \begin{subfigure}[b]{0.23\linewidth}
       \includegraphics[width=\linewidth]{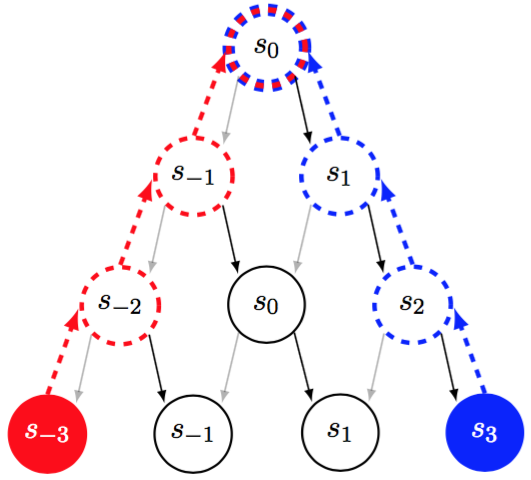}
       \caption{RL+deep explore}
       \label{fig: rl deep}
   \end{subfigure}
   \caption{Planning, learning and exploration in RL.}
   \label{fig: deep explore}
\end{figure}
}

\subsection{Testing for deep exploration}
\vspace{-2mm}

\begin{figure}[h]
 \centering
 \vspace{-4mm}
 \includegraphics[width=0.66\linewidth]{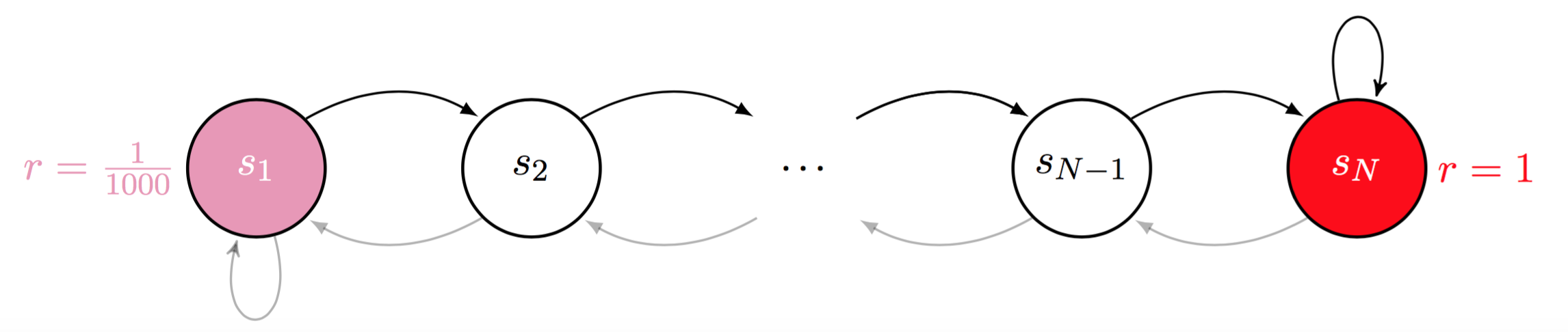}
 \vspace{-2mm}
 \caption{\small Scalable environments that requires deep exploration.}
 \vspace{-2mm}
 \label{fig: N chain example}
\end{figure}

We now present a series of didactic computational experiments designed to highlight the need for deep exploration.
These environments can be described by chains of length $N > 3$ in Figure \ref{fig: N chain example}.
Each episode of interaction lasts $N+9$ steps after which point the agent resets to the initial state $s_2$.
These are toy problems intended to be expository rather than entirely realistic.
Balancing a well known and mildly successful strategy versus an unknown, but potentially more rewarding, approach can emerge in many practical applications.

These environments may be described by a finite tabular MDP.
However, we consider algorithms which interact with the MDP only through raw pixel features.
We consider two feature mappings $\phi_{\rm 1hot}(s_t) := (\Ind\{x = s_t\})$ and $\phi_{\rm therm}(s_t) := (\Ind\{x \le s_t\})$ in $\{0,1\}^N$.
We present results for $\phi_{\rm therm}$, which worked better for all DQN variants due to better generalization, but the difference was relatively small - see Appendix \ref{app: marshmallow test}.
Thompson DQN is the same as bootstrapped DQN, but resamples every timestep.
Ensemble DQN uses the same architecture as bootstrapped DQN, but with an ensemble policy.

\begin{figure}[!h]
 \centering
 \vspace{-4mm}
 \includegraphics[width=0.9\linewidth]{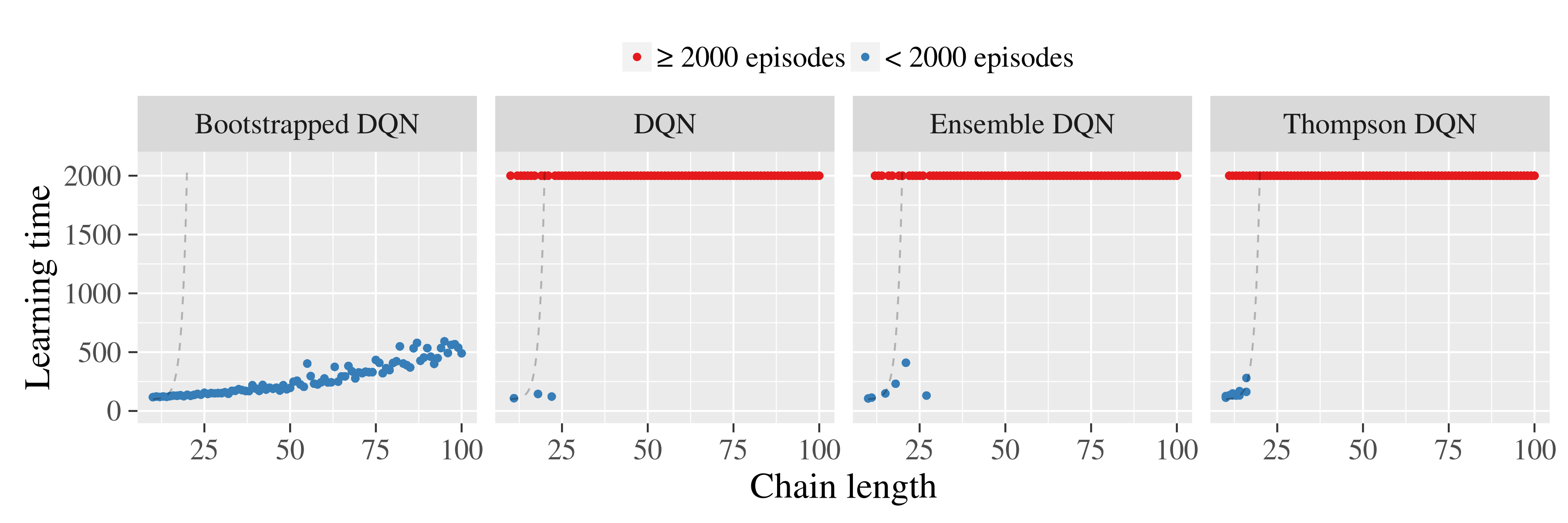}
 \vspace{-3mm}
 \caption{Only Bootstrapped DQN demonstrates deep exploration.}
 \vspace{-3mm}
 \label{fig: chain performance}
\end{figure}

We say that the algorithm has successfully learned the optimal policy when it has successfully completed one hundred episodes with optimal reward of $10$.
For each chain length, we ran each learning algorithm for 2000 episodes across three seeds.
We plot the median time to learn in Figure \ref{fig: chain performance}, together with a conservative lower bound of $99 + 2^{N-11}$ on the expected time to learn for any shallow exploration strategy \cite{osband2014generalization}.
Only bootstrapped DQN demonstrates a graceful scaling to long chains which require deep exploration.

\subsection{How does bootstrapped DQN drive deep exploration?}
\label{sec: how deep}
\vspace{-2mm}

Bootstrapped DQN explores in a manner similar to the provably-efficient algorithm PSRL \cite{Osband2013} but it uses a bootstrapped neural network to approximate a posterior sample for the value.
Unlike PSRL, bootstrapped DQN directly samples a value function and so does not require further planning steps.
This algorithm is similar to RLSVI, which is also provably-efficient \cite{osband2014generalization}, but with a neural network instead of linear value function and bootstrap instead of Gaussian sampling.
The analysis for the linear setting suggests that this nonlinear approach will work well so long as the distribution $\{Q^1,..,Q^K\}$ remains \textit{stochastically optimistic}~\cite{osband2014generalization}, or at least as spread out as the ``correct'' posterior.

Bootstrapped DQN relies upon random initialization of the network weights as a prior to induce diversity.
Surprisingly, we found this initial diversity was enough to maintain diverse generalization to new and unseen states for large and deep neural networks.
This is effective for our experimental setting, but will not work in all situations.
In general it may be necessary to maintain some more rigorous notion of ``prior'', potentially through the use of artificial prior data to maintain diversity \cite{osband2015bootstrapped}.
One potential explanation for the efficacy of simple random initialization is that unlike supervised learning or bandits, where all networks fit the same data, each of our $Q^k$ heads has a unique target network.
This, together with stochastic minibatch and flexible nonlinear representations, means that even small differences at initialization may become \textit{bigger} as they refit to unique TD errors.

Bootstrapped DQN does \textit{not} require that any single network $Q^k$ is initialized to the correct policy of ``right'' at every step, which would be exponentially unlikely for large chains $N$.
For the algorithm to be successful in this example we only require that the networks generalize in a diverse way to the actions they have never chosen in the states they have not visited very often.
Imagine that, in the example above, the network has made it as far as state $\tilde{N} < N$, but never observed the action right $a=2$.
As long as one head $k$ imagines $Q(\tilde{N}, 2) > Q(\tilde{N}, 2)$ then TD bootstrapping can propagate this signal back to $s=1$ through the target network to drive deep exploration.
The expected time for these estimates at $n$ to propagate to at least one head grows gracefully in $n$, even for relatively small $K$, as our experiments show.
We expand upon this intuition with a video designed to highlight \textit{how} bootstrapped DQN demonstrates deep exploration \url{https://youtu.be/e3KuV_d0EMk}.
We present further evaluation on a difficult stochastic MDP in Appendix \ref{app: marshmallow test}.

\vspace{-1mm}
\section{Arcade Learning Environment}
\label{sec: atari}
\vspace{-3mm}

We now evaluate our algorithm across 49 Atari games on the Arcade Learning Environment \cite{bellemare2012arcade}.
Importantly, and unlike the experiments in Section \ref{sec: marshmallow}, these domains are not specifically designed to showcase our algorithm.
In fact, many Atari games are structured so that small rewards always indicate part of an optimal policy.
This may be crucial for the strong performance observed by dithering strategies\footnote{By contrast, imagine that the agent received a small immediate reward for dying; dithering strategies would be hopeless at solving this problem, just like Section \ref{sec: marshmallow}.}.
We find that exploration via bootstrapped DQN produces significant gains versus $\epsilon$-greedy in this setting.
Bootstrapped DQN reaches peak performance roughly similar to DQN.
However, our improved exploration mean we reach human performance on average 30\% faster across all games.
This translates to significantly improved cumulative rewards through learning.

We follow the setup of \cite{van2015deep} for our network architecture and benchmark our performance against their algorithm.
Our network structure is identical to the convolutional structure of DQN \cite{mnih2015human} except we split 10 separate bootstrap heads after the convolutional layer as per Figure \ref{fig: shared convnet}.
Recently, several authors have provided architectural and algorithmic improvements to DDQN \cite{wang2015dueling,schaul2015prioritized}.
We do not compare our results to these since their advances are orthogonal to our concern and could easily be incorporated to our bootstrapped DQN design.
Full details of our experimental set up are available in Appendix \ref{app: atari}.

\subsection{Implementing bootstrapped DQN at scale}
\vspace{-2mm}
We now examine how to generate online bootstrap samples for DQN in a computationally efficient manner.
We focus on three key questions: how many heads do we need, how should we pass gradients to the shared network and how should we bootstrap data online?
We make significant compromises in order to maintain computational cost comparable to DQN.

Figure \ref{fig: nBoot} presents the cumulative reward of bootstrapped DQN on the game Breakout, for different number of heads $K$.
More heads leads to faster learning, but even a small number of heads captures most of the benefits of bootstrapped DQN.
We choose $K=10$.

{\footnotesize
\begin{figure}[!h]
   \centering
   \vspace{-2mm}
   \begin{subfigure}[b]{0.4\linewidth}
       \includegraphics[width=\linewidth]{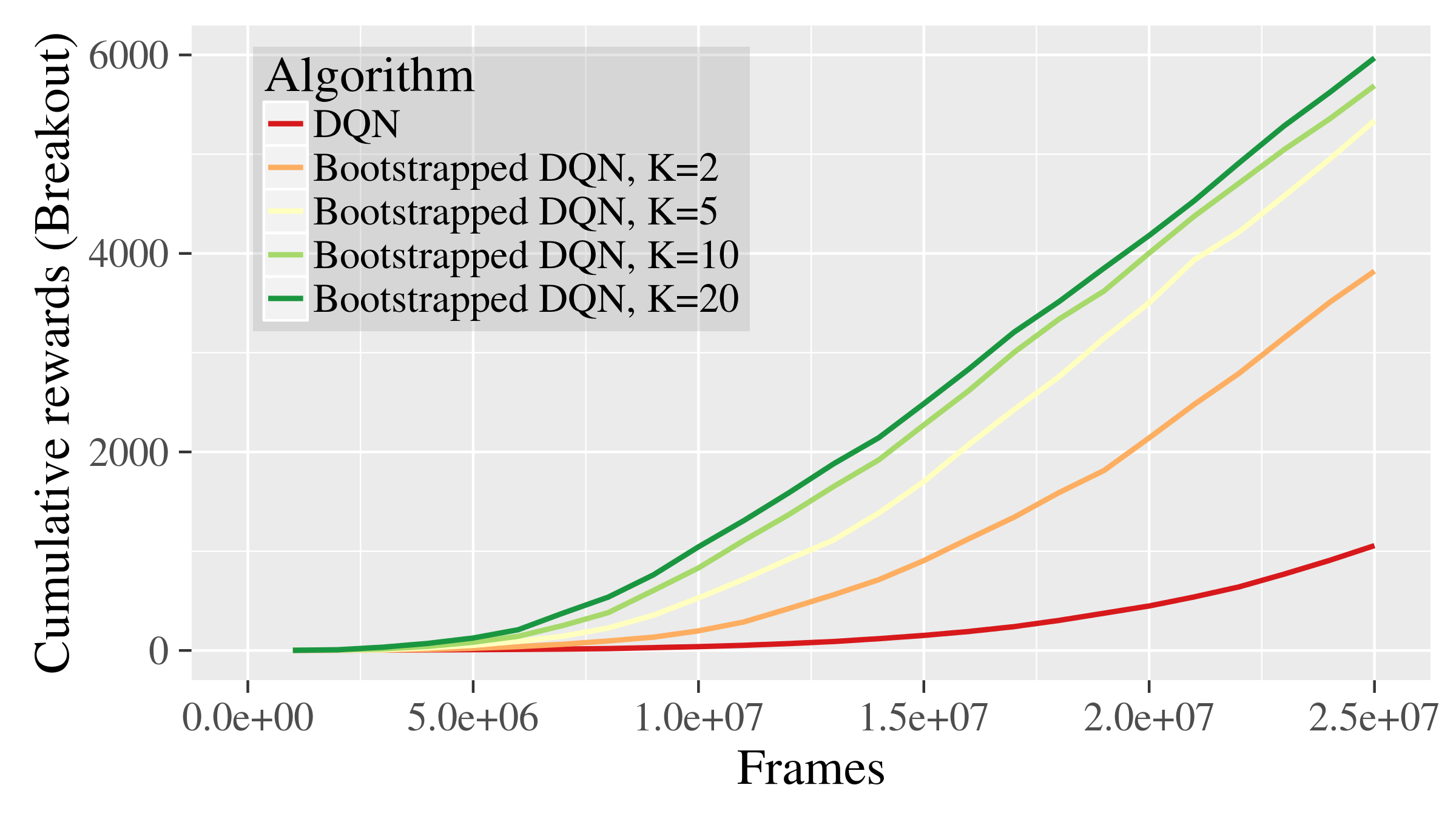}
       \vspace{-6mm}
       \caption{Number of bootstrap heads $K$.}
       \label{fig: nBoot}
   \end{subfigure}
   \hspace{4mm}
   \begin{subfigure}[b]{0.4\linewidth}
       \includegraphics[width=\linewidth]{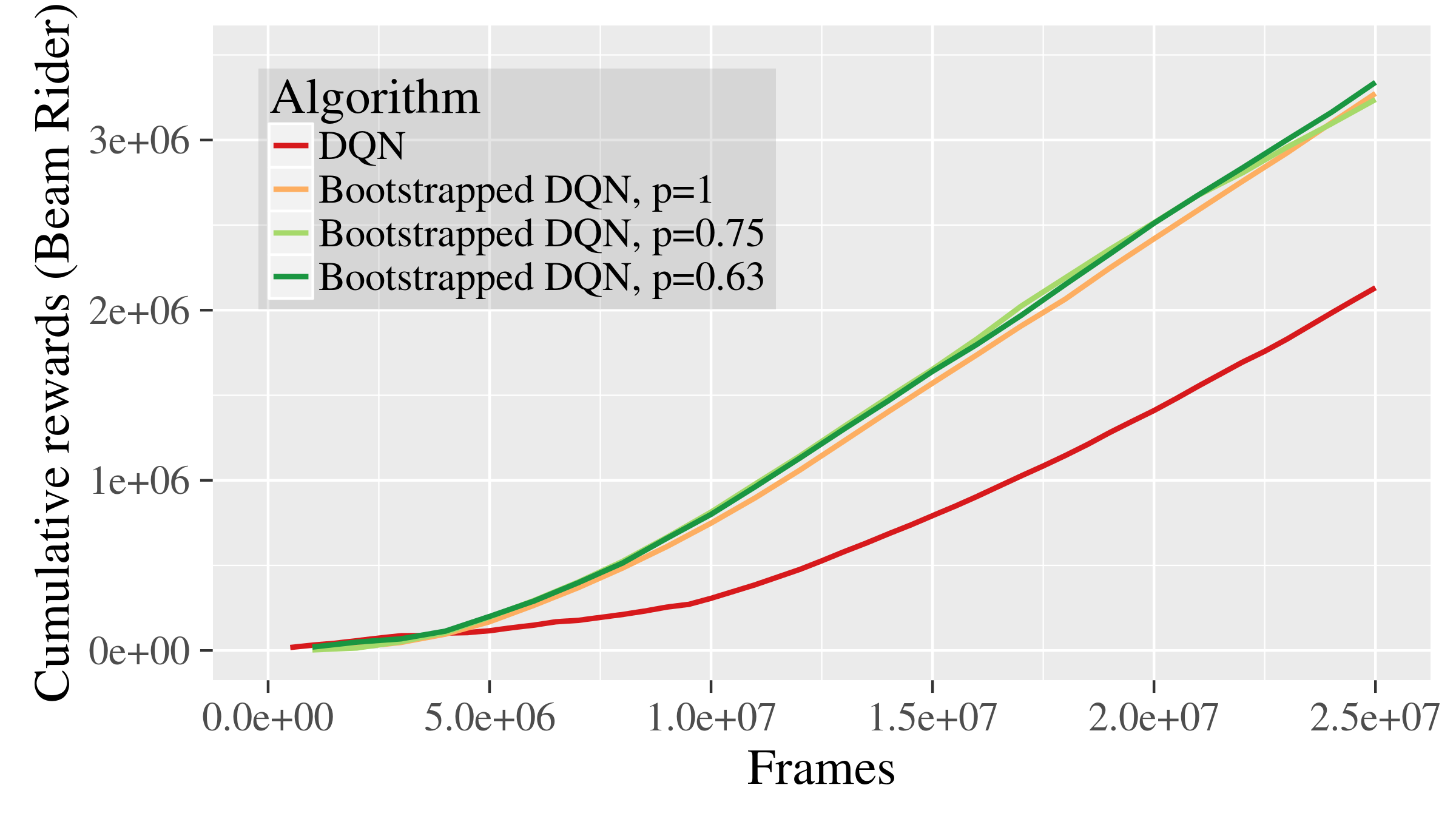}
       \vspace{-6mm}
       \caption{Probability of data sharing $p$.}
       \label{fig: probShare2}
   \end{subfigure}
   \vspace{-2mm}
   \caption{Examining the sensitivities of bootstrapped DQN.}
   \vspace{-2mm}
   \label{fig: learning sens}
\end{figure}
}

The shared network architecture allows us to train this combined network via backpropagation.
Feeding $K$ network heads to the shared convolutional network effectively increases the learning rate for this portion of the network.
In some games, this leads to premature and sub-optimal convergence.
We found the best final scores by normalizing the gradients by $1/K$, but this also leads to slower early learning.
See Appendix \ref{app: atari} for more details.

{\medmuskip=0mu
\thinmuskip=0mu
\thickmuskip=0mu
To implement an online bootstrap we use an independent Bernoulli mask $w_1, .., w_K \sim {\rm Ber}(p)$ for each head in each episode\footnote{$p=0.5$ is double-or-nothing bootstrap \cite{owen2012bootstrapping}, $p=1$ is ensemble with no bootstrapping at all.}.
These flags are stored in the memory replay buffer and identify which heads are trained on which data.
However, when trained using a shared minibatch the algorithm will also require an effective $1/p$ more iterations; this is undesirable computationally.
Surprisingly, we found the algorithm performed similarly irrespective of $p$ and all outperformed DQN, as shown in Figure \ref{fig: probShare2}.
This is strange and we discuss this phenomenon in Appendix \ref{app: atari}.
However, in light of this empirical observation for Atari, we chose $p=1$ to save on minibatch passes.
As a result bootstrapped DQN runs at similar computational speed to vanilla DQN on identical hardware\footnote{Our implementation {\medmuskip=0mu
\thinmuskip=0mu
\thickmuskip=0mu$K=10, \ p=1$} ran with less than a $20$\% increase on wall-time versus DQN.}.
}

\vspace{-1mm}
\subsection{Efficient exploration in Atari}
\begin{figure}[!h]
\centering
    \vspace{-2mm}
    \includegraphics[width=\linewidth]{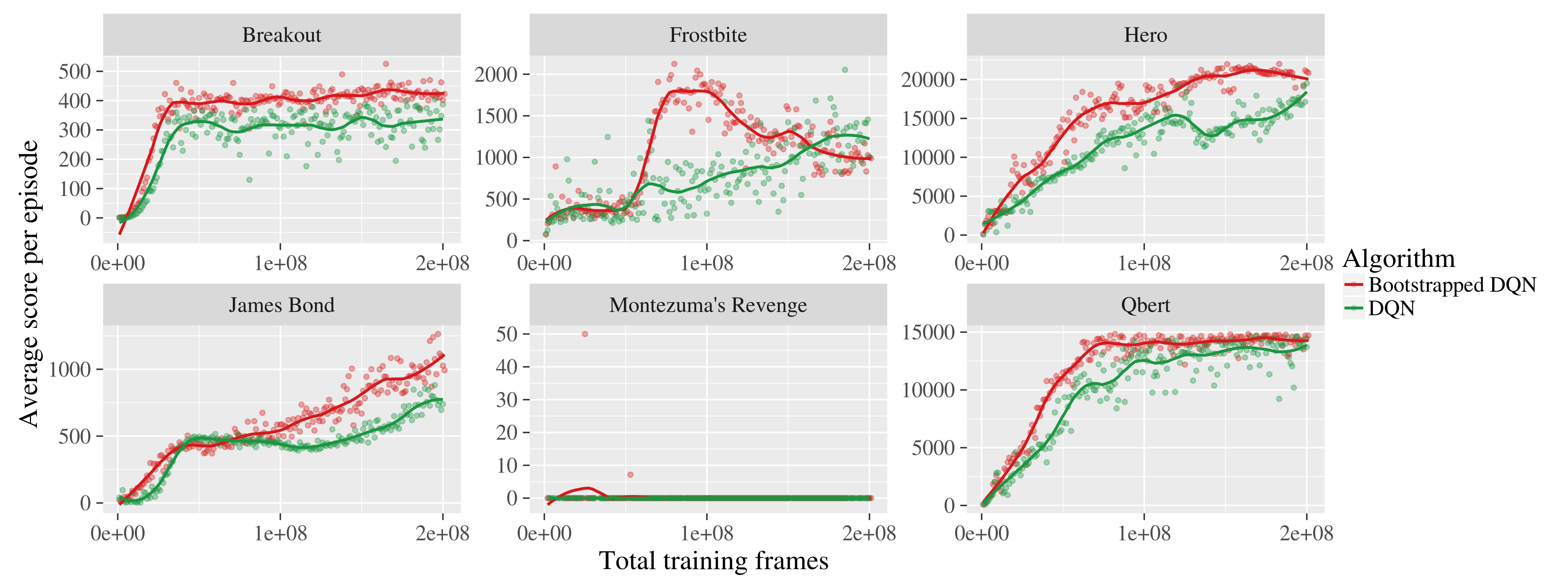}
    \vspace{-7mm}
   \caption{\small Bootstrapped DQN drives more efficient exploration.}
   \vspace{-2mm}
   \label{fig: bootstrap learn}
\end{figure}

\vspace{-1mm}
We find that Bootstrapped DQN drives efficient exploration in several Atari games.
For the same amount of game experience, bootstrapped DQN generally outperforms DQN with $\epsilon$-greedy exploration.
Figure \ref{fig: bootstrap learn} demonstrates this effect for a diverse selection of games.

\begin{figure}[h!]
 \centering
 \vspace{-3mm}
 \includegraphics[width=0.85\linewidth]{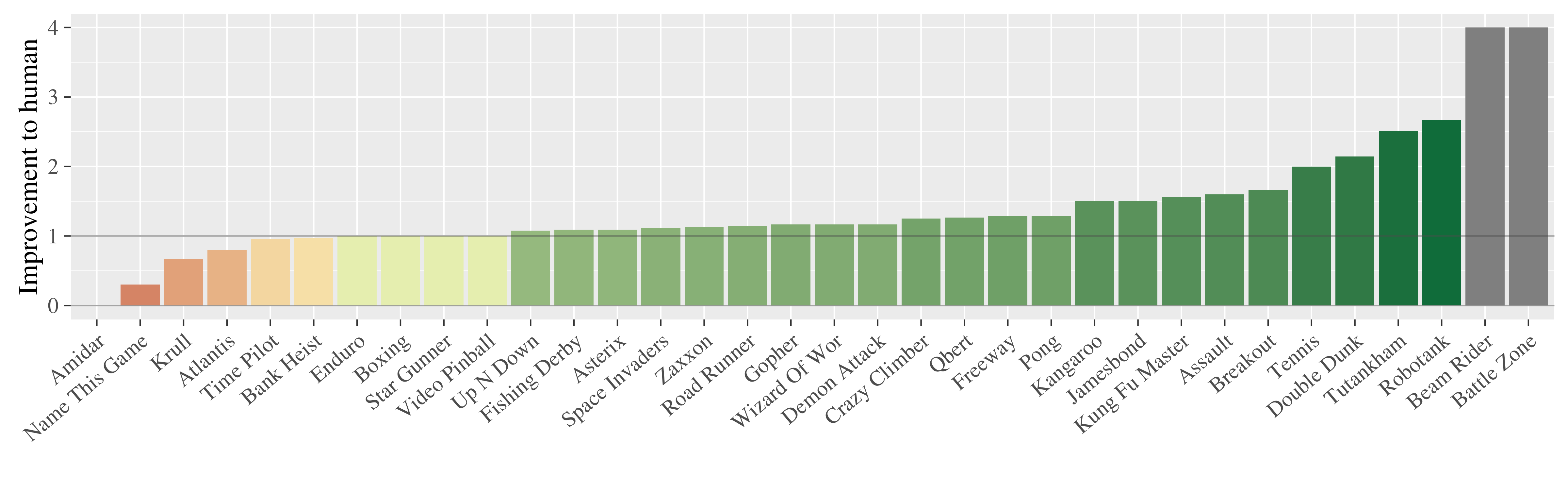}
 \vspace{-7mm}
 \caption{\small Bootstrapped DQN reaches human performance faster than DQN.}
 \vspace{-2mm}
 \label{fig: time to human}
\end{figure}

On games where DQN performs well, bootstrapped DQN typically performs better.
Bootstrapped DQN does not reach human performance on Amidar (DQN does) but does on Beam Rider and Battle Zone (DQN does not).
To summarize this improvement in learning time we consider the number of frames required to reach human performance.
If bootstrapped DQN reaches human performance in $1/x$ frames of DQN we say it has improved by $x$.
Figure \ref{fig: time to human} shows that Bootstrapped DQN typically reaches human performance significantly faster.

On most games where DQN does not reach human performance, bootstrapped DQN does not solve the problem by itself.
On some challenging Atari games where deep exploration is conjectured to be important \cite{van2015deep} our results are not entirely successful, but still promising.
In Frostbite, bootstrapped DQN reaches the second level much faster than DQN but network instabilities cause the performance to crash.
In Montezuma's Revenge, bootstrapped DQN reaches the first key after 20m frames (DQN never observes a reward even after 200m frames) but does not properly learn from this experience\footnote{An improved training method, such as prioritized replay \cite{schaul2015prioritized} may help solve this problem.}.
Our results suggest that improved exploration may help to solve these remaining games, but also highlight the importance of other problems like network instability, reward clipping and temporally extended rewards.

\vspace{-2mm}
\subsection{Overall performance}
\vspace{-2mm}

\begin{figure}[h!]
 \centering
 \vspace{-3mm}
 \includegraphics[width=0.85\linewidth]{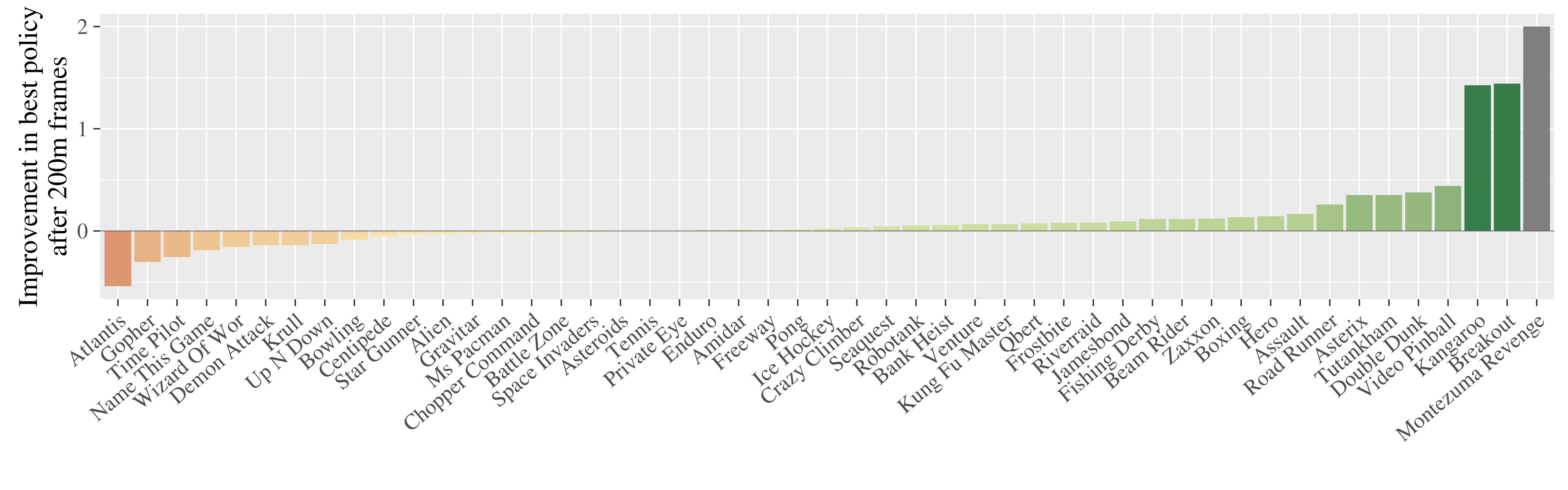}
 \vspace{-7mm}
 \caption{\small Bootstrapped DQN typically improves upon the best policy.}
 \vspace{-2mm}
 \label{fig: final score}
\end{figure}

\begin{figure}[h!]
 \centering
 \vspace{-3mm}
 \includegraphics[width=0.85\linewidth]{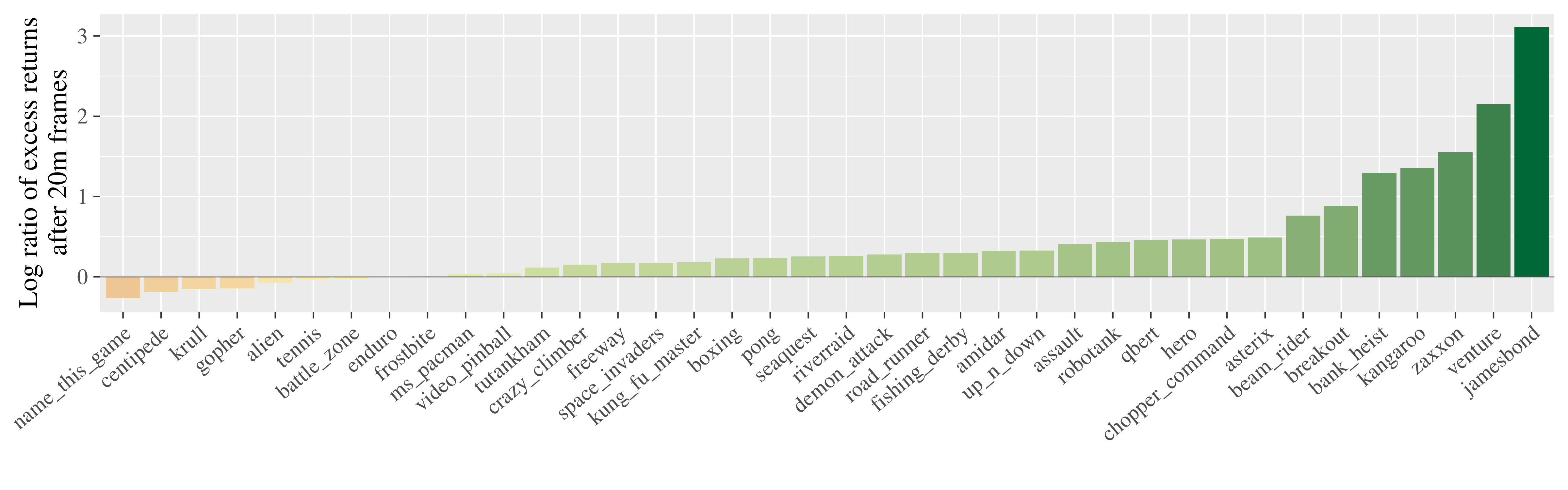}
 \vspace{-7mm}
 \caption{\small Bootstrapped DQN improves cumulative rewards by orders of magnitude.}
 \vspace{-4mm}
 \label{fig: cum rewards}
\end{figure}

Bootstrapped DQN is able to learn much faster than DQN.
Figure \ref{fig: final score} shows that bootstrapped DQN also improves upon the final score across most games.
However, the real benefits to \textit{efficient} exploration mean that bootstrapped DQN outperforms DQN by orders of magnitude in terms of the \textit{cumulative} rewards through learning (Figure \ref{fig: cum rewards}.
In both figures we normalize performance relative to a fully random policy.
The most similar work to ours presents several other approaches to improved exploration in Atari \cite{stadie2015incentivizing} they optimize for AUC-20, a normalized version of the cumulative returns after 20m frames.
According to their metric, averaged across the 14 games they consider, we improve upon both base DQN (0.29) and their best method (0.37) to obtain 0.62 via bootstrapped DQN.
We present these results together with results tables across all 49 games in Appendix \ref{app: final results}.

\subsection{Visualizing bootstrapped DQN}
\vspace{-2mm}

We now present some more insight to how bootstrapped DQN drives deep exploration in Atari.
In each game, although each head $Q^1, ..,Q^{10}$ learns a high scoring policy, the policies they find are quite distinct.
In the video  \url{https://youtu.be/Zm2KoT82O_M} we show the evolution of these policies simultaneously for several games.
Although each head performs well, they each follow a unique policy.
By contrast, $\epsilon$-greedy strategies are almost indistinguishable for small values of $\epsilon$ and totally ineffectual for larger values.
We believe that this deep exploration is key to improved learning, since diverse experiences allow for better generalization.

Disregarding exploration, bootstrapped DQN may be beneficial as a purely exploitative policy.
We can combine all the heads into a single ensemble policy, for example by choosing the action with the most votes across heads.
This approach might have several benefits.
First, we find that the ensemble policy can often outperform any individual policy.
Second, the distribution of votes across heads to give a measure of the uncertainty in the optimal policy.
Unlike vanilla DQN, bootstrapped DQN can know what it doesn't know.
In an application where executing a poorly-understood action is dangerous this could be crucial.
In the video \url{https://youtu.be/0jvEcC5JvGY} we visualize this ensemble policy across several games.
We find that the uncertainty in this policy is surprisingly interpretable: all heads agree at clearly crucial decision points, but remain diverse at other less important steps.

\vspace{-2mm}
\section{Closing remarks}
\vspace{-2mm}
In this paper we present bootstrapped DQN as an algorithm for efficient reinforcement learning in complex environments.
We demonstrate that the bootstrap can produce useful uncertainty estimates for deep neural networks.
Bootstrapped DQN is computationally tractable and also naturally scalable to massive parallel systems.
We believe that, beyond our specific implementation, randomized value functions represent a promising alternative to dithering for exploration.
Bootstrapped DQN practically combines efficient generalization with exploration for complex nonlinear value functions.

\newpage

{
\small
\bibliographystyle{plain}
\bibliography{reference}
}

\newpage
\appendix

\begin{center}
\textbf{\Large APPENDICES}
\end{center}

\section{Uncertainty for neural networks}
\label{app: uncertainty nn}

\begin{figure}[h!]
 \centering
 \includegraphics[width=0.9 \linewidth]{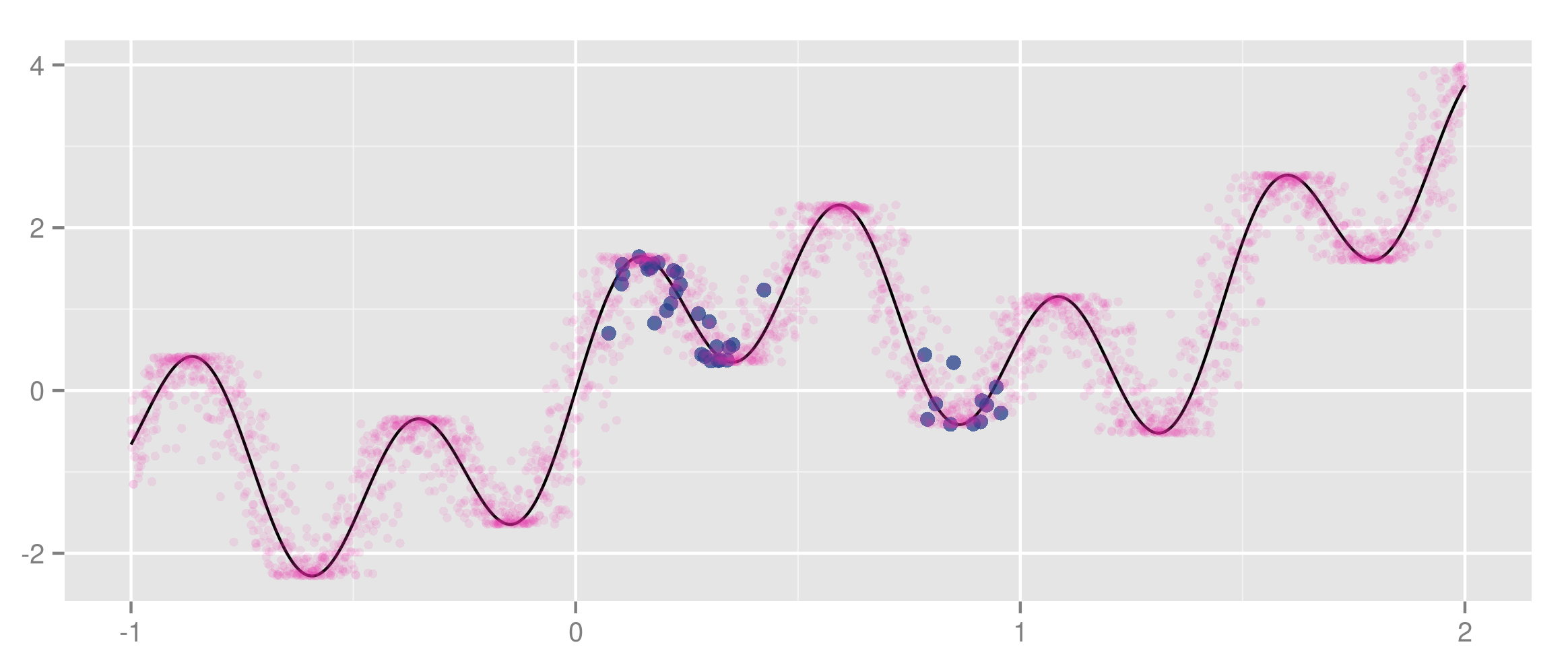}
 \caption{\small Underlying generating distribution. All our algorithms receive the same blue data. Pink points represent other samples, the mean function is shown in black.}
 \label{fig: true uncertainty}
\end{figure}

In this appendix we discuss some of the experimental setup to qualitatively evaluate uncertainty methods for deep neural networks.
To do this, we generated twenty noisy regression pairs $x_i, y_i$ with:
$$ y_i = x_i + sin(\alpha (x_i + w_i)) + sin(\beta (x_i + w_i)) + w_i $$
where $x_i$ are drawn uniformly from $(0,0.6) \cup (0.8, 1)$ and $w_i \sim N(\mu=0, \sigma^2=0.03^2)$.
We set $\alpha=4$ and $\beta=13$.
None of these numerical choices were important except to represent a highly nonlinear function with lots of noise and several clear regions where we should be uncertain.
We present the regression data together with an indication of the generating distribution in Figure \ref{fig: true uncertainty}.

Interestingly, we did not find that using dropout produced satisfying confidence intervals for this task.
We present one example of this dropout posterior estimate in Figure \ref{fig: dropout uncertainty}.

{\footnotesize
\begin{figure}[!h]
   \centering
   \vspace{-2mm}
   \begin{subfigure}[b]{0.45\linewidth}
    \includegraphics[width=0.9 \linewidth]{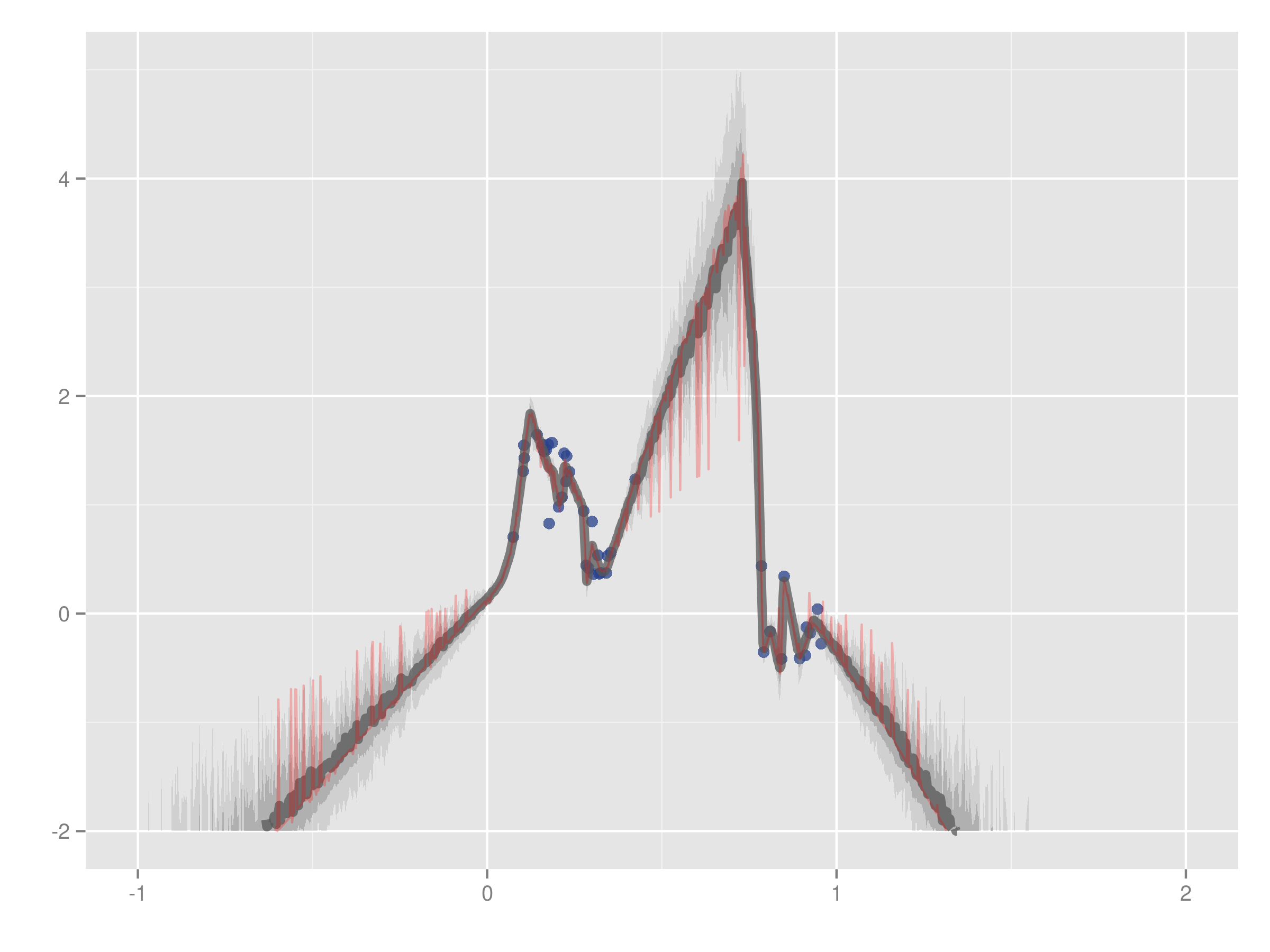}
    \caption{\small Dropout gives strange uncertainty estimates.}
       \label{fig: dropout uncertainty}
   \end{subfigure}
   \hspace{4mm}
   \begin{subfigure}[b]{0.45\linewidth}
 \includegraphics[width=0.8 \linewidth]{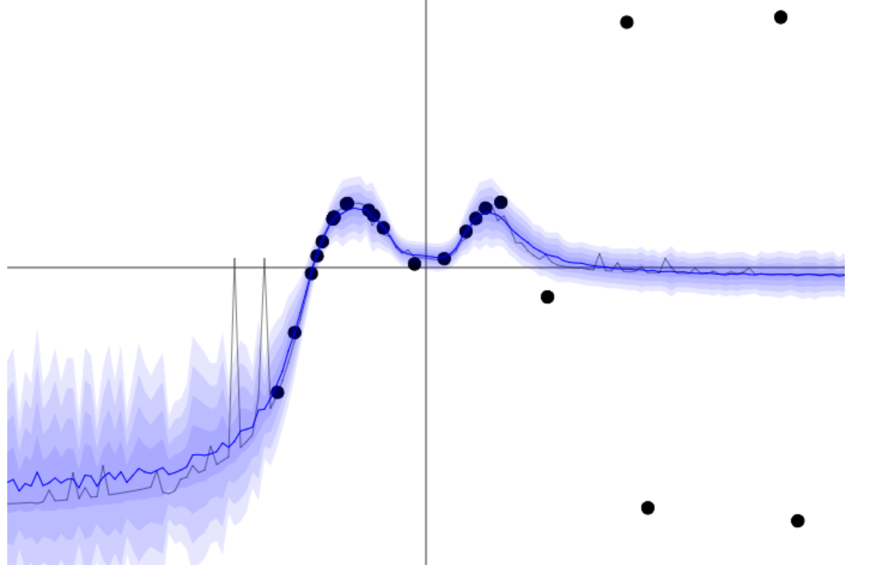}
 \caption{\small Screenshot from accompanying web demo to \cite{gal2015dropout}. Dropout converges with high certainty to the mean value.}
 \label{fig: dropout yarin}
   \end{subfigure}
   \vspace{-2mm}
   \caption{Comparing the bootstrap to dropout uncertainty for neural nets.}
   \vspace{-2mm}
   \label{fig: boot}
\end{figure}
}

These results are unsatisfactory for several reasons.
First, the network extrapolates the mean posterior far outside the range of any actual data for $x=0.75$.
We believe this is because dropout only perturbs locally from a single neural network fit, unlike bootstrap.
Second, the posterior samples from the dropout approximation are very spiky and do not look like any sensible posterior sample.
Third, the network collapses to almost zero uncertainty in regions with data.

We spent some time altering our dropout scheme to fix this effect, which might be undesirable for stochastic domains and we believed might be an artefact of our implementation.
However, after further thought we believe this to be an effect which you would expect for dropout posterior approximations.
In Figure \ref{fig: dropout yarin} we present a didactic example taken from the author's website \cite{gal2015dropout}.

On the right hand side of the plot we generate noisy data with wildly different values.
Training a neural network using MSE criterion means that the network will surely converge to the mean of the noisy data.
Any dropout samples remain highly concentrated around this mean.
By contrast, bootstrapped neural networks may include different subsets of this noisy data and so may produce a more intuitive uncertainty estimates for our settings.
Note this isn't necessarily a failure of dropout to approximate a Gaussian process posterior, but this artefact could be shared by any homoskedastic posterior.
The authors of \cite{gal2015dropout} propose a heteroskedastic variant which can help, but does not address the fundamental issue that for large networks trained to convergence all dropout samples may converge to every single datapoint... even the outliers.

In this paper we focus on the bootstrap approach to uncertainty for neural networks.
We like its simplicity, connections to established statistical methodology and empirical good performance.
However, the key insights of this paper is the use of deep exploration via randomized value functions.
This is compatible with any approximate posterior estimator for deep neural networks.
We believe that this area of uncertainty estimates for neural networks remains an important area of research in its own right.

Bootstrapped uncertainty estimates for the Q-value functions have another crucial advantage over dropout which does not appear in the supervised problem.
Unlike random dropout masks trained against random target networks, our implementation of bootstrap DQN trains against its own \textit{temporally consistent} target network.
This means that our bootstrap estimates (in the sense of \cite{efron1982jackknife}), are able to ``bootstrap'' (in the TD sense of \cite{Sutton1998}) on their own estimates of the long run value.
This is important to quantify the long run uncertainty over Q and drive deep exploration.

\section{Bootstrapped DQN implementation}
\label{app: boot dqn algorithm}

Algorithm~\ref{alg:full} gives a full description of Bootstrapped DQN.
It captures two modes of operation where either $k$ neural networks
are used to estimate the $Q_k$-value functions, or
where one neural network with $k$ heads is used to estimate $k$ $Q$-value functions.
In both cases, as this is largely a parameterisation issue, we denote the
value function networks as $Q$, where $Q_k$ is output of the $k$th network or
the $k$th head.

A core idea to the full bootstrapped DQN algorithm is the bootstrap mask $m_t$.
The mask $m_t$ decides, for each value function $Q_k$, whether or not it should train upon the experience
generated at step $t$.
In its simplest form $m_t$ is a binary vector of length $K$, masking out or including each value function for
training on that time step of experience (i.e., should it receive gradients from the
corresponding $(s_t, a_t, r_{t+1}, s_{t+1}, m_t)$ tuple).
The masking distribution $M$ is responsible for generating each $m_t$.
For example, when $M$ yields $m_t$ whose components are independently drawn from a
bernoulli distribution with parameter $0.5$ then this corresponds to the
double-or-nothing bootstrap \cite{owen2012bootstrapping}.
On the other hand, if $M$ yields a mask $m_t$ with all ones, then the
algorithm reduces to an ensemble method.
Poisson masks $M_t[k] \sim {\rm Poi}(1)$ provides the most natural parallel with the standard non-parameteric boostrap since ${\rm Bin}(N, 1/N) \rightarrow {\rm Poi}(1)$ as $N \rightarrow \infty$.
Exponential masks $M_t[k] \sim {\rm Exp}(1)$ closely resemble the standard Bayesian nonparametric posterior of a Dirichlet process \cite{osband2015bootstrapped}.

\begin{algorithm}
    \caption{Bootstrapped DQN \label{alg:full}}
    \begin{algorithmic}[1]
        \State \textbf{Input:} Value function networks $Q$ with $K$ outputs $\{Q_k\}_{k=1}^K$. Masking distribution $M$.
\State{Let $B$ be a replay buffer storing experience for training.}
\For{each episode}
	\State{Obtain initial state from environment $s_0$}
	\State{Pick a value function to act using $k \sim \text{Uniform}\{1,\dots,K\}$}
	\For{step $t = 1,\ldots$ until end of episode}
		\State{Pick an action according to $a_t \in \arg\max_a Q_k(s_t, a)$}
		\State{Receive state $s_{t+1}$ and reward $r_t$ from environment, having taking action $a_t$}
		\State{Sample bootstrap mask $m_t \sim M$}
		\State{Add $(s_t, a_t, r_{t+1}, s_{t+1}, m_t)$ to replay buffer $B$}
	\EndFor
\EndFor
    \end{algorithmic}
\end{algorithm}

Periodically, the replay buffer is played back to update the parameters of the value function network $Q$.
The gradients of the $k$th value function $Q_k$ for the $t$th tuple in the replay buffer $B$, $g_t^k$ is:
\begin{align}
g_t^k &= m_t^k(y^Q_t - Q_k(s_t, a_t; \theta))\nabla_\theta Q_k(s_t, a_t; \theta)
\end{align}
where $y^Q_t$ is given by \eqref{eq:yqt}.
Note that the mask $m_t^k$ modulates the gradient, giving rise to the bootstrap behaviour.

\section{Experiments for deep exploration}
\label{app: marshmallow test}


\subsection{Bootstrap methodology}

A naive implementation of bootstrapped DQN builds up $K$ complete networks with $K$ distinct memory buffers.
This method is parallelizable up to many machines, however we wanted to produce an algorithm that was efficient even on a single machine.
To do this, we implemented the bootstrap heads in a single larger network, like Figure \ref{fig: shared convnet} but without any shared network.
We implement bootstrap by masking each episode of data according to $w_1,..,w_K \sim {\rm Ber}(p)$.

\begin{figure}[h!]
 \centering
 \includegraphics[width=0.75 \linewidth]{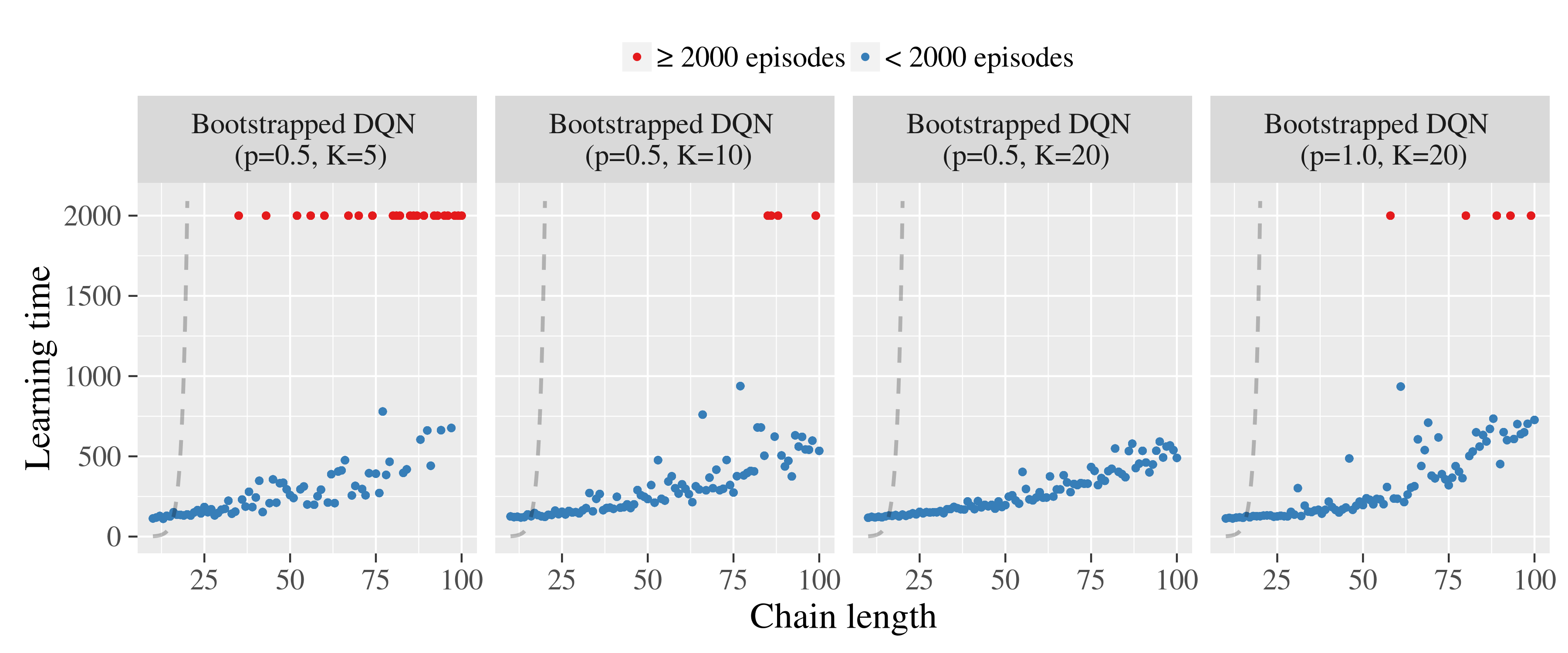}
 \caption{\small Bootstrapped DQN performs well even with small number of bootstrap heads $K$ or high probability of sharing $p$.}
 \label{fig: deep explore app}
\end{figure}

In Figure \ref{fig: deep explore app} we demonstrate that bootstrapped DQN can implement deep exploration even with relatively small values of $K$.
However, the results are more robust and scalable with larger $K$.
We run our experiments on the example from Figure \ref{fig: N chain example}.
Surprisingly, this method is even effective with $p=1$ and complete data sharing between heads.
This degenerate full sharing of information turns out to be remarkably efficient for training large and deep neural networks.
We discuss this phenomenon more in Appendix \ref{app: atari}.

\begin{figure}[h!]
 \centering
 \includegraphics[width=0.75 \linewidth]{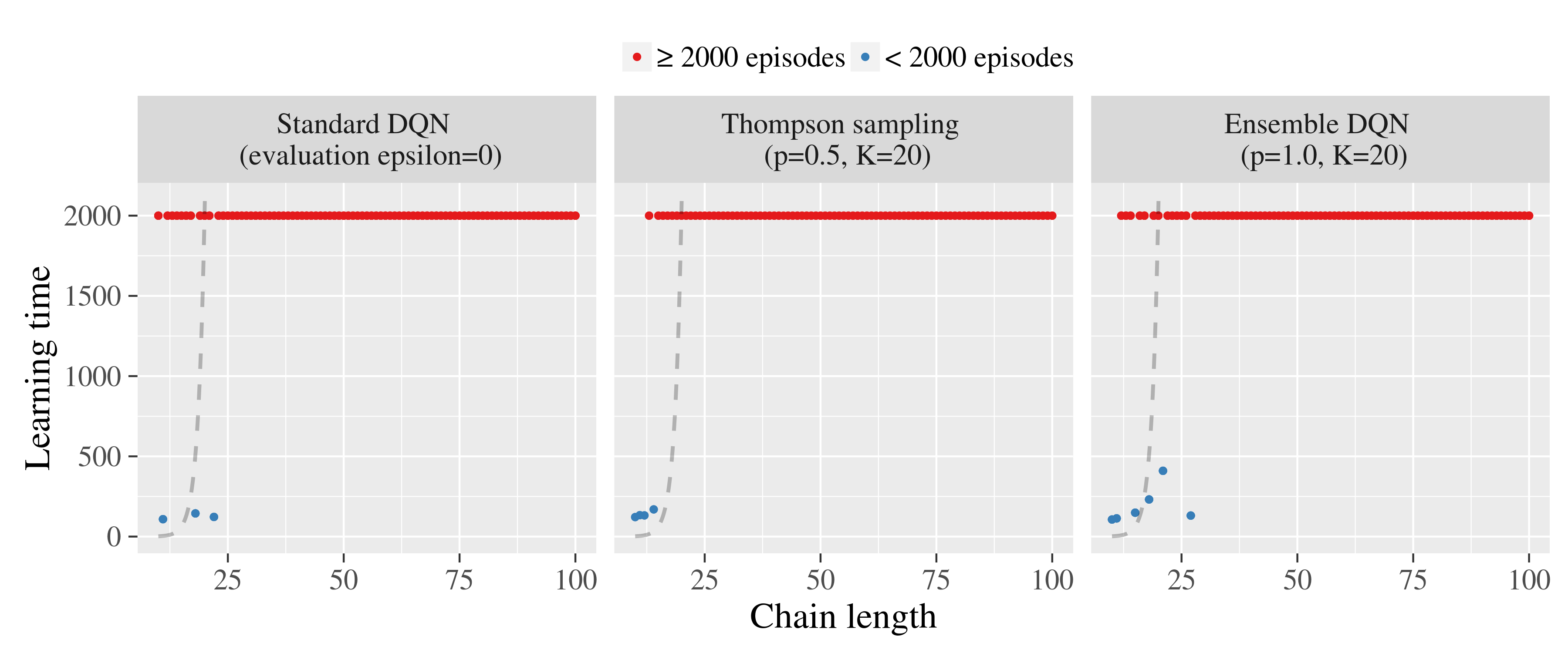}
 \caption{\small Shallow exploration methods do not work.}
 \label{fig: bad explore app}
\end{figure}

Generating good estimates for uncertainty is not enough for efficient exploration.
In Figure \ref{fig: bad explore app} we see that other methods trained with the same network architecture are totally ineffective at implementing deep exploration.
The $\epsilon$-greedy policy follows just one $Q$-value estimate.
We allow this policy to be evaluated without dithering.
The ensemble policy is trained exactly as per bootstrapped DQN except at each stage the algorithm follows the policy which is majority vote of the bootstrap heads.
Thompson sampling is the same as bootstrapped DQN except a new head is sampled every timestep, rather than every episode.

We can see that only bootstrapped DQN demonstrates efficient and deep exploration in this domain.

\subsection{A difficult stochastic MDP}
\vspace{-1mm}

\begin{figure}[h!]
 \vspace{-3mm}
 \centering
 \includegraphics[width=0.95\linewidth]{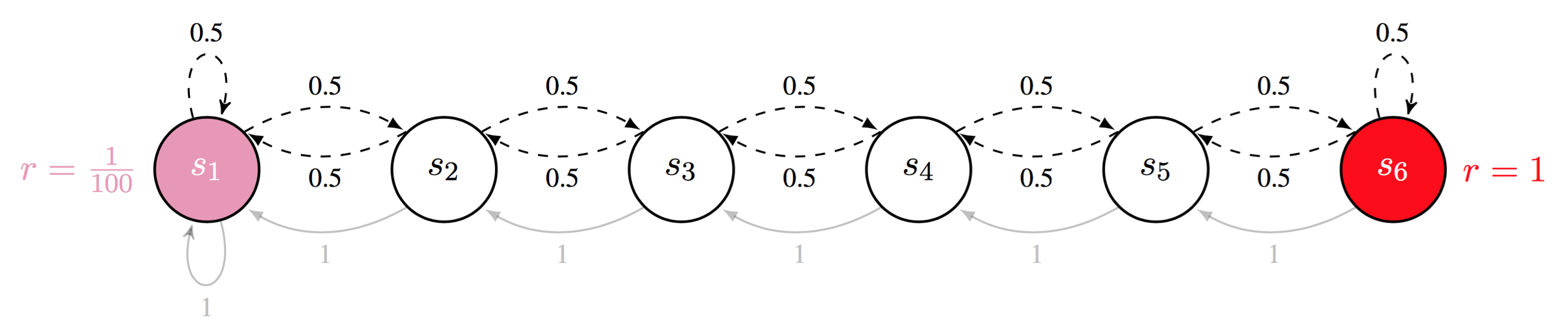}
 \vspace{-3mm}
 \caption{A stochastic MDP that requires deep exploration.}
 \vspace{-2mm}
 \label{fig: 6 chain example}
\end{figure}

Figure \ref{fig: chain performance} shows that bootstrapped DQN can implement effective (and deep) exploration where similar deep RL architectures fail.
However, since the underlying system is a small and finite MDP there may be several other simpler strategies which would also solve this problem.
We will now consider a difficult variant of this chain system with significant stochastic noise in transitions as depicted in Figure \ref{fig: 6 chain example}.
Action ``left'' deterministically moves the agent left, but action ``right'' is only successful 50\% of the time and otherwise also moves left.
The agent interacts with the MDP in episodes of length $15$ and begins each episode at $s_1$.
Once again the optimal policy is to head right.

Bootstrapped DQN is unique amongst scalable approaches to efficient exploration with deep RL in stochastic domains.
For benchmark performance we implement three algorithms which, unlike bootstrapped DQN, will receive the true tabular representation for the MDP.
These algorithms are based on three state of the art approaches to exploration via dithering ($\epsilon$-greedy), optimism \cite{Jaksch2010} and posterior sampling \cite{Osband2013}.
We discuss the choice of these benchmarks in Appendix \ref{app: marshmallow test}.

{\footnotesize
\begin{figure}[!h]
   \centering
   \vspace{-2mm}
   \begin{subfigure}[b]{0.45\linewidth}
 \includegraphics[width=0.99\linewidth]{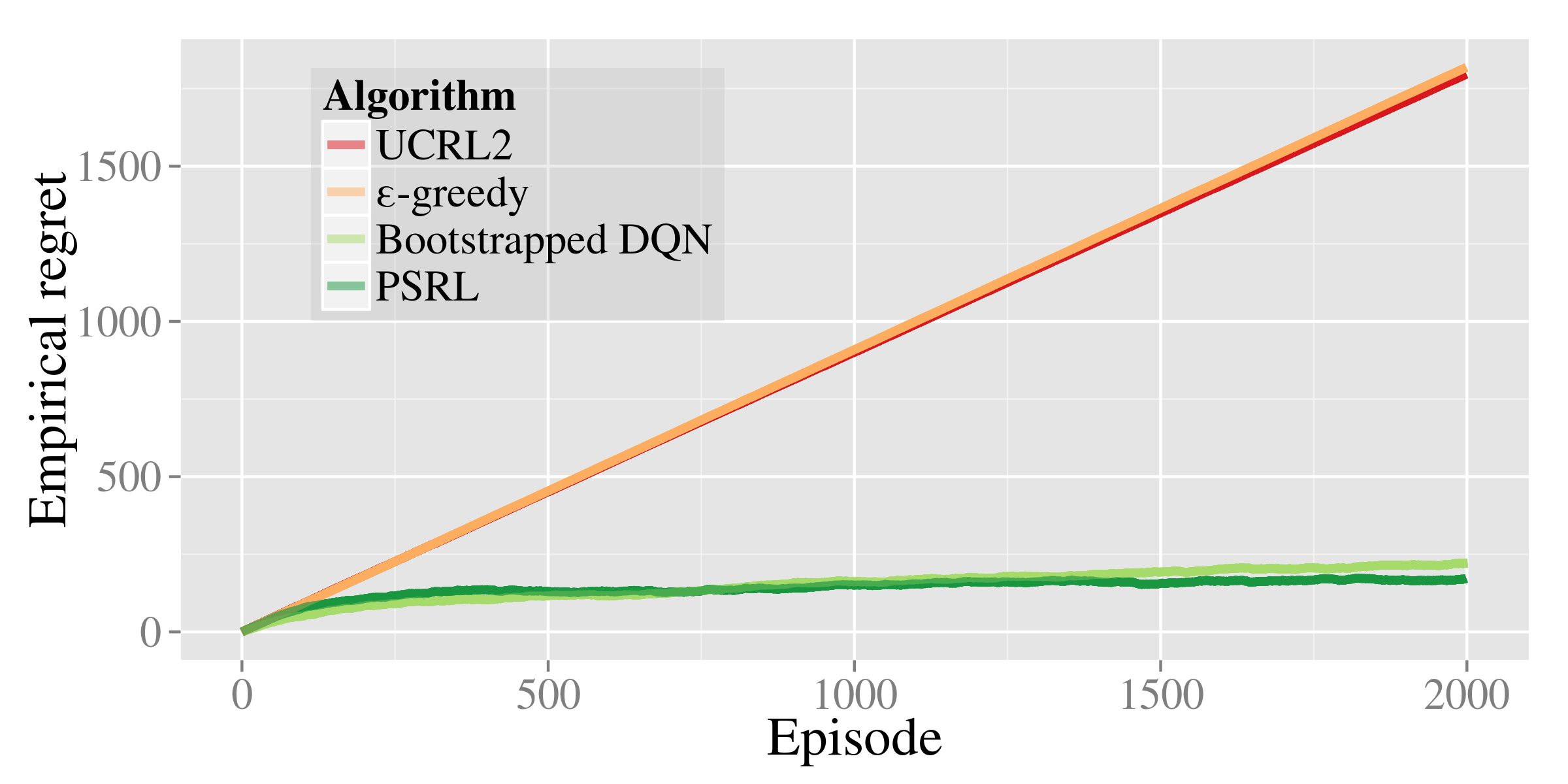}
 \vspace{-3mm}
 \caption{\small Bootstrapped DQN matches efficient tabular RL.}
 \label{fig: chain 50 regret}
   \end{subfigure}
   \hspace{4mm}
   \begin{subfigure}[b]{0.45\linewidth}
 \includegraphics[width=0.99\linewidth]{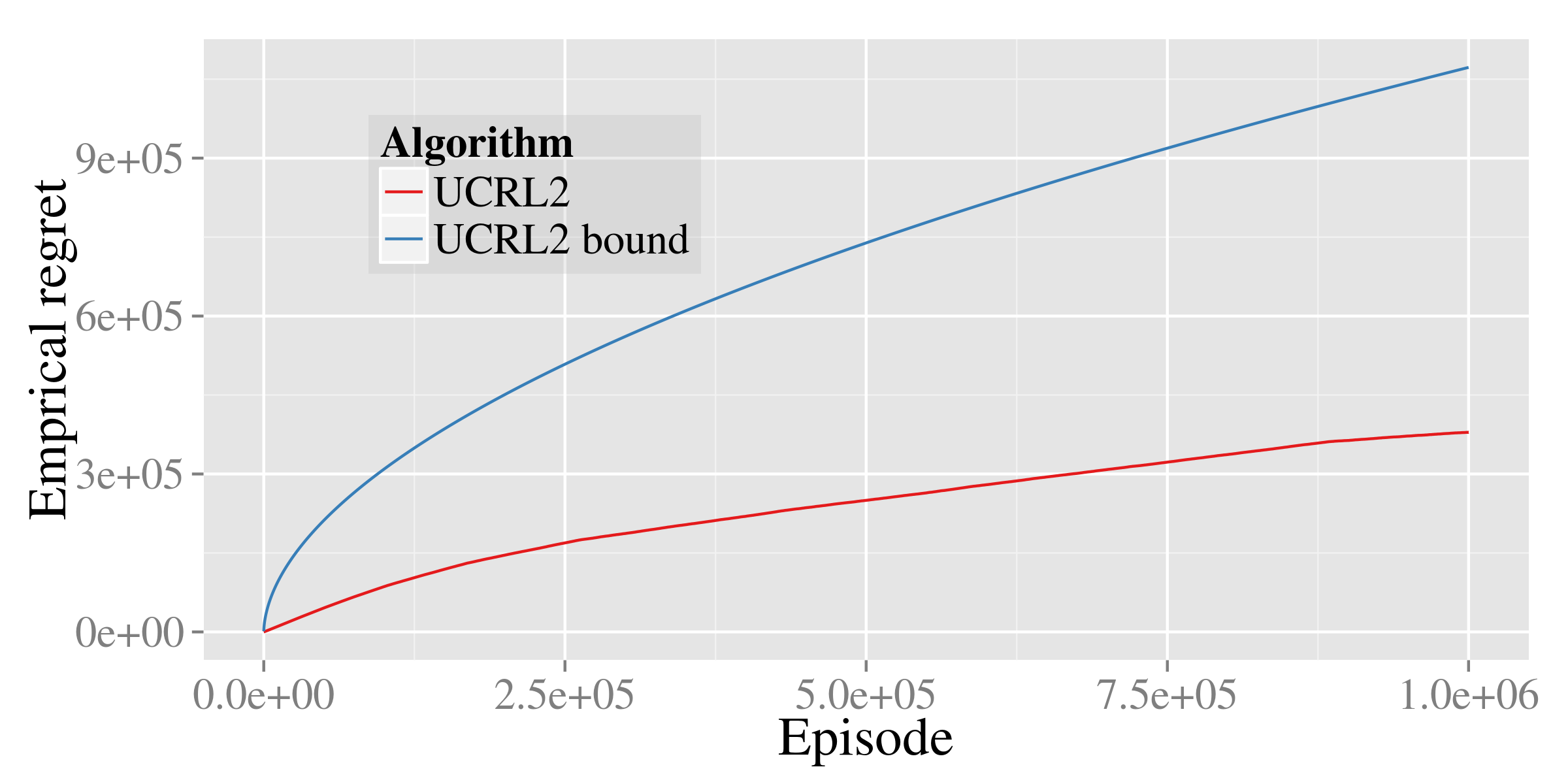}
 \caption{\small The regret bounds for UCRL2 are near-optimal in $\tilde{O}(\cdot)$, but they are still not very practical.}
 \label{fig: ucrl2 long}
   \end{subfigure}
   \vspace{-2mm}
   \caption{Learning and regret bounds on a stochastic MDP.}
   \vspace{-2mm}
   \label{fig: stochastic MDP}
\end{figure}
}

In Figure \ref{fig: chain 50 regret} we present the empirical regret of each algorithm averaged over 10 seeds over the first two thousand episodes.
The empirical regret is the cumulative difference between the expected rewards of the optimal policy and the realized rewards of each algorithm.
We find that bootstrapped DQN achieves similar performance to state of the art efficient exploration schemes such as PSRL even without prior knowledge of the tabular MDP structure and in noisy environments.

Most telling is how much better bootstrapped DQN does than the state of the art optimistic algorithm UCRL2.
Although Figure \ref{fig: chain 50 regret} seems to suggest UCRL2 incurs linear regret, actually it follows its bounds $\tilde{O} ( S \sqrt{AT} )$ \cite{Jaksch2010} where $S$ is the number of states and $A$ is the number of actions.

For the example in Figure \ref{fig: 6 chain example} we attempted to display our performance compared to several benchmark tabula rasa approaches to exploration.
There are many other algorithms we could have considered, but for a short paper we chose to focus against the most common approach ($\epsilon$-greedy) the pre-eminent optimistic approach (UCRL2) and posterior sampling (PSRL).

Other common heuristic approaches, such as optimistic initialization for Q-learning can be tuned to work well on this domain, however the precise parameters are sensitive to the underlying MDP\footnote{Further, it is difficult to extend the idea of optimistic initialization with function generalization, especially for deep neural networks.}.
To make a general-purpose version of this heuristic essentially leads to optimistic algorithms.
Since UCRL2 is originally designed for infinite-horizon MDPs, we use the natural adaptation of this algorithm, which has state of the art guarantees in finite horizon MDPs as well \cite{dann2015sample}.

Figure \ref{fig: chain 50 regret} displays the empirical regret of these algorithms together with bootstrapped DQN on the example from Figure \ref{fig: 6 chain example}.
It is somewhat disconcerting that UCRL2 appears to incur linear regret, but it is proven to satisfy near-optimal regret bounds.
Actually, as we show in Figure \ref{fig: ucrl2 long}, the algorithm produces regret which scales very similarly to its established bounds \cite{Jaksch2010}.
Similarly, even for this tiny problem size, the recent analysis that proves a near optimal sample complexity in fixed horizon problems \cite{dann2015sample} only guarantees that we will have fewer than $10^{10}$ $\epsilon=1$ suboptimal episodes.
While these bounds may be acceptable in worst case $\tilde{O}(\cdot)$ scaling, they are not of much practical use.

\subsection{One-hot features}

\begin{figure}[h!]
 \centering
 \includegraphics[width=0.5\linewidth]{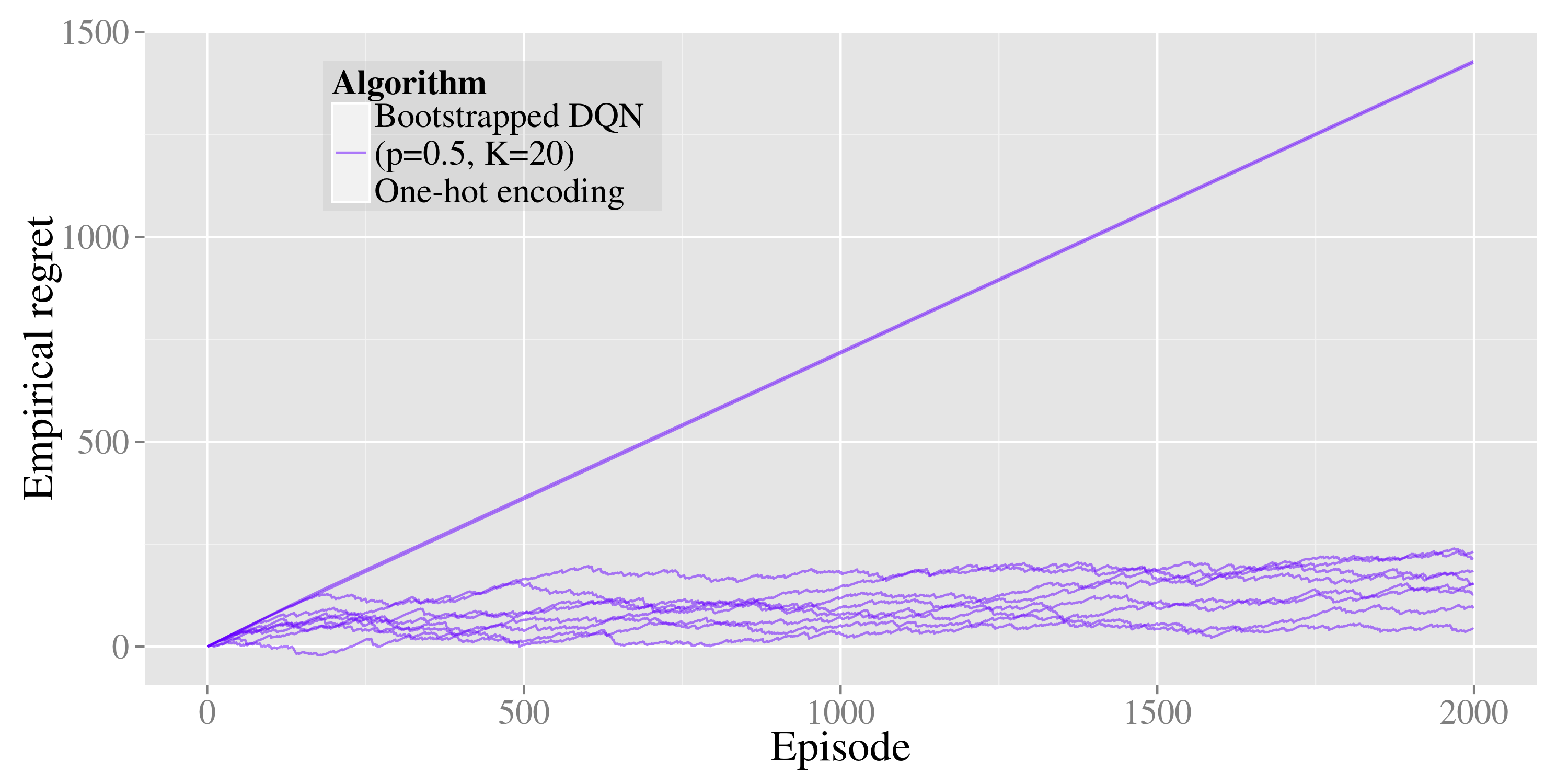}
 \caption{\small Bootstrapped DQN also performs well with one-hot features, but learning is less robust.}
 \label{fig: one-hot}
\end{figure}

In Figure \ref{fig: one-hot} we include the mean performance of bootstrapped DQN with one-hot feature encodings.
We found that, using these features, bootstrapped DQN learned the optimal policy for most seeds, but was somewhat less robust than the thermometer encoding.
Two out of ten seeds failed to learn the optimal policy within 2000 episodes, this is presented in Figure \ref{fig: one-hot}.

\section{Experiments for Atari}
\label{app: atari}

\subsection{Experimental setup}
We use the same 49 Atari games as \cite{mnih2015human} for our experiments.
Each step of the agent corresponds to four steps of the emulator, where the same action is repeated, the reward values of the agents are clipped between -1 and 1 for stability.
We evaluate our agents and report performance based upon the raw scores.

The convolutional part of the network used is identical to the one used in \cite{mnih2015human}.
The input to the network is 4x84x84 tensor with a rescaled, grayscale version of the last four observations.
The first convolutional (conv) layer has 32 filters of size 8 with a stride of 4.
The second conv layer has 64 filters of size 4 with stride 2.
The last conv layer has 64 filters of size 3.
We split the network beyond the final layer into $K=10$ distinct heads, each one is fully connected and identical to the single head of DQN \cite{mnih2015human}.
This consists of a fully connected layer to 512 units followed by another fully connected layer to the Q-Values for each action.
The fully connected layers all use Rectified Linear Units(ReLU) as a non-linearity.
We normalize gradients $1/K$ that flow from each head.

We trained the networks with RMSProp with a momentum of 0.95 and a learning rate of 0.00025 as in \cite{mnih2015human}.
The discount was set to $\gamma = 0.99$, the number of steps between target updates was set to $\tau = 10000$ steps.
We trained the agents for a total of 50m steps per game, which corresponds to 200m frames.
The agents were every 1m frames, for evaluation in bootstrapped DQN we use an ensemble voting policy.
The experience replay contains the 1m most recent transitions.
We update the network every $4$ steps by randomly sampling a minibatch of $32$ transitions from the replay buffer to use the exact same minibatch schedule as DQN.
For training we used an $\epsilon$-greedy policy with $\epsilon$ being annealed linearly from $1$ to $0.01$ over the first 1m timesteps.

\subsection{Gradient normalization in bootstrap heads}
\label{app: grad norm}
Most literature in deep RL for Atari focuses on learning the best single evaluation policy, with particular attention to whether this above or below human performance \cite{mnih2015human}.
This is unusual for the RL literature, which typically focuses upon cumulative or final performance.

\begin{figure}[h!]
 \centering
 \includegraphics[width=0.9 \linewidth]{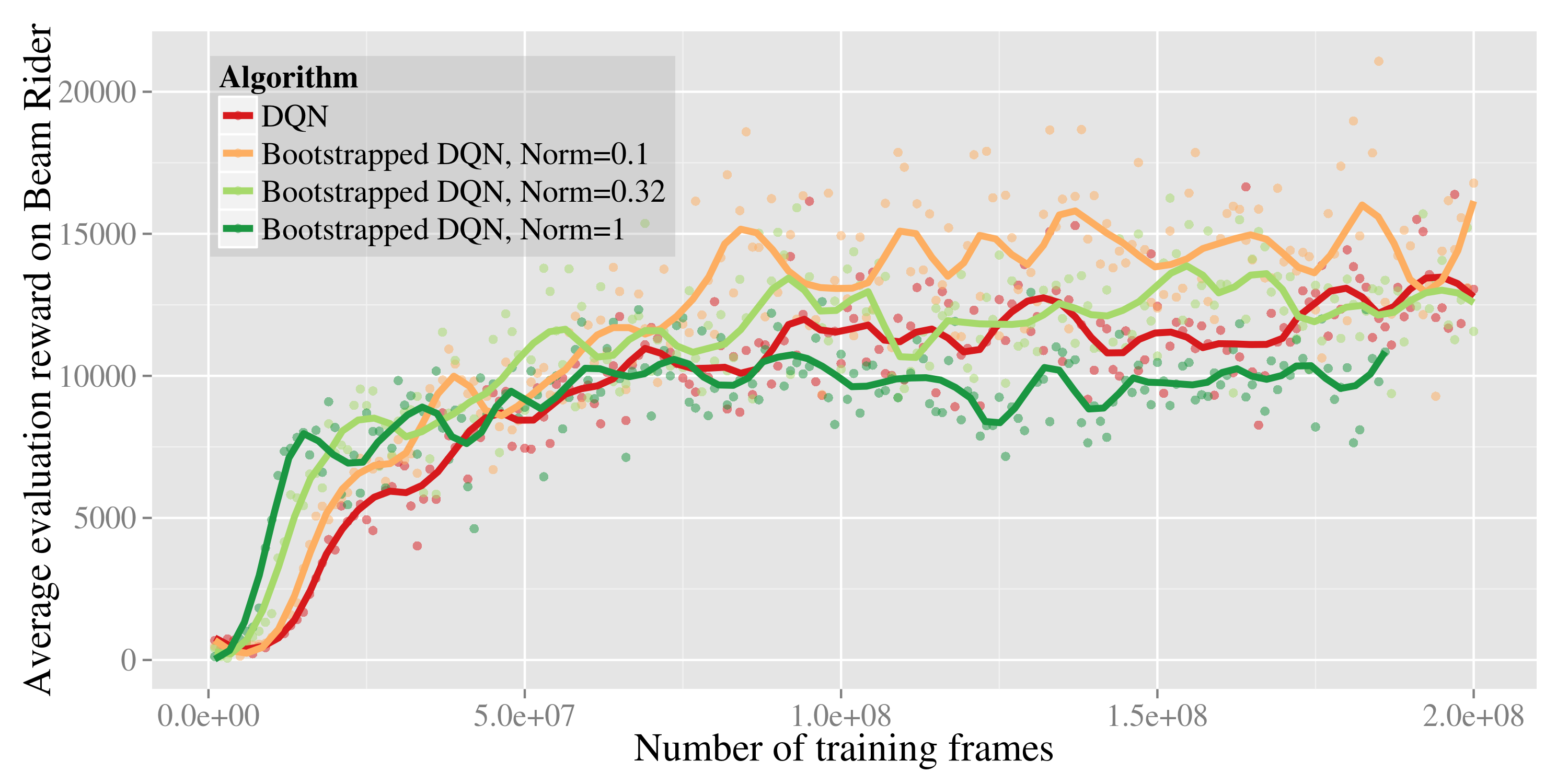}
 \caption{\small Normalization fights premature convergence.}
 \label{fig: norm plateau}
\end{figure}

Bootstrapped DQN makes significant improvements to the cumulative rewards of DQN on Atari, as we display in Figure \ref{fig: cum rewards}, while the peak performance is much more
We found that using bootstrapped DQN without gradient normalization on each head typically learned even faster than our implementation with rescaling $1/K$, but it was somewhat prone to premature and suboptimal convergence.
We present an example of this phenomenon in Figure \ref{fig: norm plateau}.

We found that, in order to better the benchmark ``best'' policies reported by DQN, it was very helpful for us to use the gradient normalization.
However, it is not entirely clear whether this represents an improvement for all settings.
In Figures \ref{fig: short norm cumulative} and \ref{fig: long norm cumulative} we present the cumulative rewards of the same algorithms on Beam Rider.

{\footnotesize
\begin{figure}[!h]
   \centering
   \vspace{-2mm}
   \begin{subfigure}[b]{0.45\linewidth}
 \includegraphics[width=0.9 \linewidth]{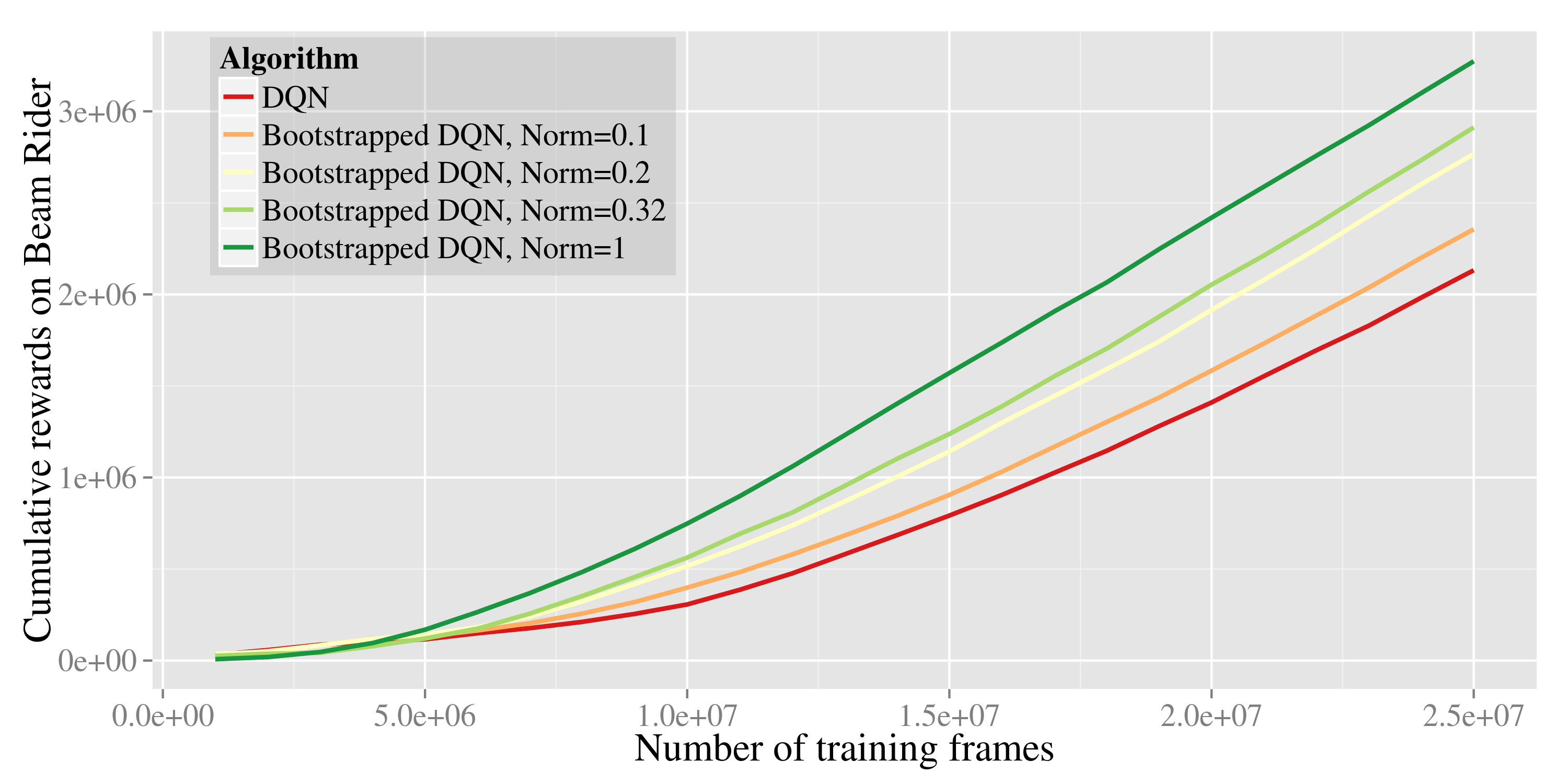}
 \caption{\small Normalization does not help cumulative rewards.}
 \label{fig: short norm cumulative}
   \end{subfigure}
   \hspace{4mm}
   \begin{subfigure}[b]{0.45\linewidth}
 \includegraphics[width=0.9 \linewidth]{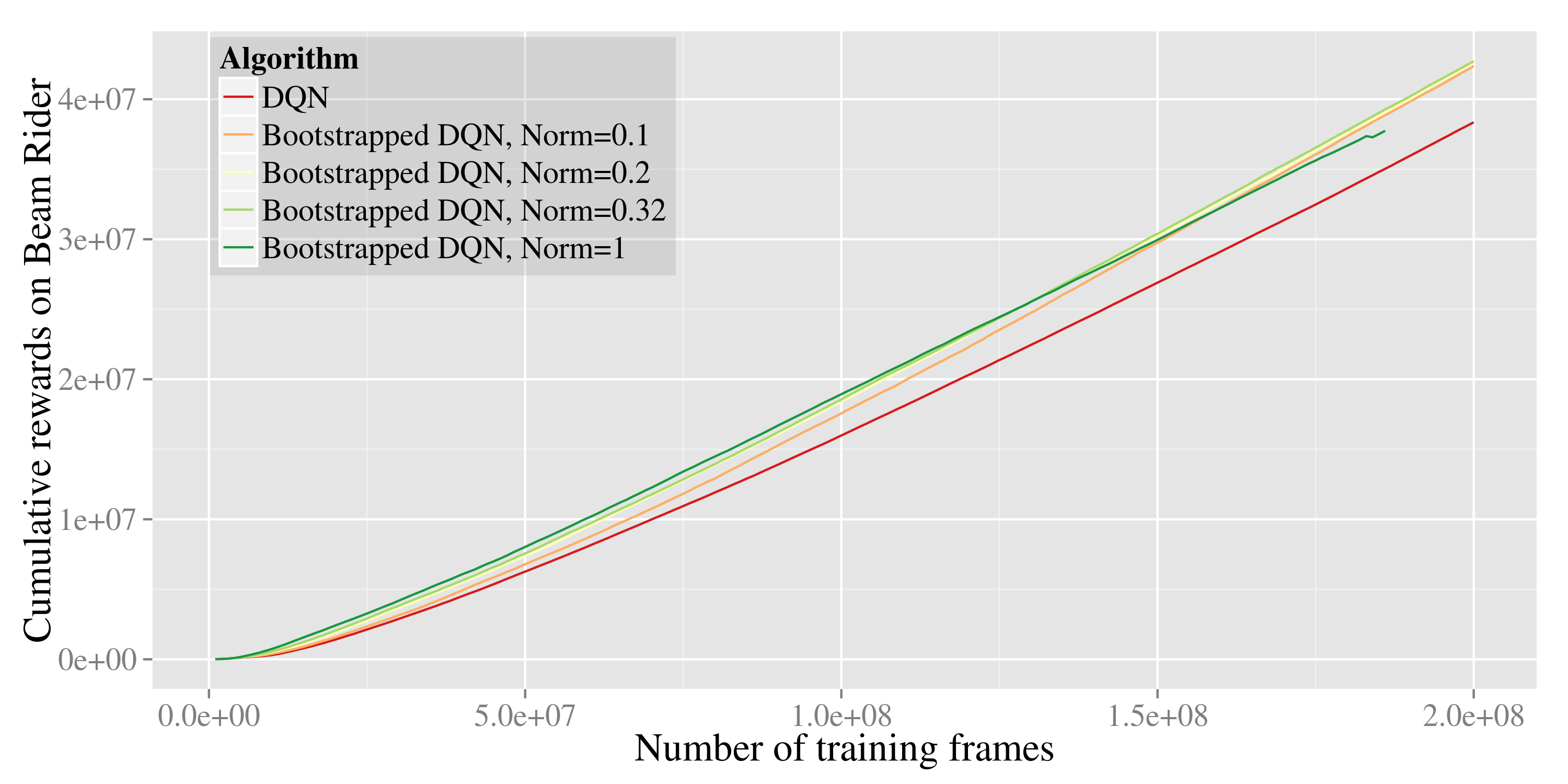}
 \caption{\small Even over 200m frames the importance of exploration dominates the effects of an inferior final policy.}
 \label{fig: long norm cumulative}
   \end{subfigure}
   \vspace{-2mm}
   \caption{Planning, learning and exploration in RL.}
   \vspace{-2mm}
   \label{fig: norm}
\end{figure}
}

Where an RL system is deployed to learn with real interactions, cumulative rewards present a better measure for performance.
In these settings the benefits of gradient normalization are less clear.
However, even with normalization $1/K$ bootstrapped DQN significantly outperforms DQN in terms of cumulative rewards.
This is reflected most clearly in Figure \ref{fig: cum rewards} and Table \ref{table: auc}.

\subsection{Sharing data in bootstrap heads}

In this setting all network heads share all the data, so they are not actually a traditional bootstrap at all.
This is different from the regression task in Section \ref{sec: uncertainty nn}, where bootstrapped data was essential to obtain meaningful uncertainty estimates.
We have several theories for why the networks maintain significant diversity even without data bootstrapping in this setting.
We build upon the intuition of Section \ref{sec: how deep}.

First, they all train on different target networks.
This means that even when facing the same $(s,a,r,s')$ datapoint this can still lead to drastically different Q-value updates.
Second, Atari is a deterministic environment, any transition observation is the unique correct datapoint for this setting.
Third, the networks are deep and initialized from different random values so they will likely find quite diverse generalization even when they agree on given data.
Finally, since all variants of DQN take many many frames to update their policy, it is likely that even using $p=0.5$ they would still populate their replay memory with identical datapoints.
This means using $p=1$ to save on minibatch passes seems like a reasonable compromise and it doesn't seem to negatively affect performance too much in this setting.
More research is needed to examine exactly where/when this data sharing is important.





\subsection{Results tables}
\label{app: final results}

In Table \ref{table: results final} the average score achieved by the agents during the most successful evaluation period, compared to human performance and a uniformly random policy.
DQN is our implementation of DQN with the hyperparameters specified above, using the double Q-Learning update.\cite{van2015deep}.
We find that peak final performance is similar under bootstrapped DQN to previous benchmarks.

To compare the benefits of exploration via bootstrapped DQN we benchmark our performance against the most similar prior work on incentivizing exploration in Atari \cite{stadie2015incentivizing}.
To do this, we compute the AUC-100 measure specified in this work.
We present these results in Table \ref{table: auc} compare to their best performing strategy as well as their implementation of DQN.
Importantly, bootstrapped DQN outperforms this prior work significantly.

{\small
\begin{center}
\begin{table}[!h]
\centering
\begin{tabular}{lccccc}
\hline
                   &   Random &   Human &   Bootstrapped DQN &      DDQN  & Nature \\
\hline
 Alien             &    227.8 &  7127.7 &             2436.6 &   \textbf{4007.7} & 3069 \\
 Amidar            &      5.8 &  1719.5 &             1272.5 &   \textbf{2138.3} & 739.5 \\
 Assault           &    222.4 &   742.0 &             \textbf{8047.1} &   6997.9 & 3359 \\
 Asterix           &    210.0 &  8503.3 &            \textbf{19713.2} &  17366.4 & 6012 \\
 Asteroids         &    719.1 & 47388.7 &             1032.0 &   \textbf{1981.4} & 1629 \\
 Atlantis          &  12850.0 & 29028.1 &           \textbf{994500.0} & 767850.0 & 85641 \\
 Bank Heist        &     14.2 &   753.1 &             \textbf{1208.0} &   1109.0 & 429.7 \\
 Battle Zone       &   2360.0 & 37187.5 &            \textbf{38666.7} &  34620.7 & 26300 \\
 Beam Rider        &    363.9 & 16926.5 &            \textbf{23429.8} &  16650.7 & 6846 \\
 Bowling           &     23.1 &   160.7 &               60.2 &     \textbf{77.9} & 42.4 \\
 Boxing            &      0.1 &    12.1 &               \textbf{93.2} &     90.2 & 71.8 \\
 Breakout          &      1.7 &    30.5 &              \textbf{855.0} &    437.0 & 401.2 \\
 Centipede         &   2090.9 & 12017.0 &             4553.5 &   4855.4 & \textbf{8309} \\
 Chopper Command   &    811.0 &  7387.8 &             4100.0 &   5019.0 & \textbf{6687} \\
 Crazy Climber     &  10780.5 & 35829.4 &           \textbf{137925.9} & 137244.4 & 114103 \\
 Demon Attack      &    152.1 &  1971.0 &            82610.0 &  \textbf{98450.0} & 9711 \\
 Double Dunk       &    -18.6 &   -16.4 &                \textbf{3.0} &     -1.8 & -18.1 \\
 Enduro            &      0.0 &   860.5 &             \textbf{1591.0} &   1496.7 & 301.8 \\
 Fishing Derby     &    -91.7 &   -38.7 &               \textbf{26.0} &     19.8 & -0.8 \\
 Freeway           &      0.0 &    29.6 &               \textbf{33.9} &     33.4 & 30.3 \\
 Frostbite         &     65.2 &  4334.7 &             2181.4 &   \textbf{2766.8} & 328.3 \\
 Gopher            &    257.6 &  2412.5 &            \textbf{17438.4} &  13815.9 & 8520 \\
 Gravitar          &    173.0 &  3351.4 &              286.1 &    \textbf{708.6} & 306.7 \\
 Hero              &   1027.0 & 30826.4 &            \textbf{21021.3} &  20974.2 & 19950 \\
 Ice Hockey        &    -11.2 &     0.9 &               \textbf{-1.3} &     -1.7 & -1.6\\
 Jamesbond         &     29.0 &   302.8 &             \textbf{1663.5} &   1120.2 & 576.7 \\
 Kangaroo          &     52.0 &  3035.0 &            \textbf{14862.5} &  14717.6 & 6740 \\
 Krull             &   1598.0 &  2665.5 &             8627.9 &   \textbf{9690.9} & 3805 \\
 Kung Fu Master    &    258.5 & 22736.3 &            \textbf{36733.3} &  36365.7 & 23270 \\
 Montezuma Revenge &      0.0 &  4753.3 &              \textbf{100.0} &      0.0 & 0 \\
 Ms Pacman         &    307.3 &  6951.6 &             2983.3 &   \textbf{3424.6} & 2311 \\
 Name This Game    &   2292.3 &  8049.0 &            11501.1 &  \textbf{11744.4} & 7257 \\
 Pong              &    -20.7 &    14.6 &               \textbf{20.9} &     \textbf{20.9} & 18.9 \\
 Private Eye       &     24.9 & 69571.3 &             \textbf{1812.5} &    158.4 & 1788 \\
 Qbert             &    163.9 & 13455.0 &            15092.7 &  \textbf{15209.7} & 10596 \\
 Riverraid         &   1338.5 & 17118.0 &            12845.0 &  \textbf{14555.1} & 8316 \\
 Road Runner       &     11.5 &  7845.0 &            \textbf{51500.0} &  49518.4 & 18257 \\
 Robotank          &      2.2 &    11.9 &               66.6 &     \textbf{70.6} & 51.6 \\
 Seaquest          &     68.4 & 42054.7 &             9083.1 &  \textbf{19183.9} & 5286 \\
 Space Invaders    &    148.0 &  1668.7 &             2893.0 &   \textbf{4715.8} & 1976 \\
 Star Gunner       &    664.0 & 10250.0 &            55725.0 &  \textbf{66091.2} & 57997\\
 Tennis            &    -23.8 &    -8.3 &                0.0 &     \textbf{11.8} & -2.5 \\
 Time Pilot        &   3568.0 &  5229.2 &             9079.4 &  \textbf{10075.8} & 5947 \\
 Tutankham         &     11.4 &   167.6 &              214.8 &    \textbf{268.0} & 186.7 \\
 Up N Down         &    533.4 & 11693.2 &            \textbf{26231.0} &  19743.5 & 8456 \\
 Venture           &      0.0 &  1187.5 &              212.5 &    239.7 & \textbf{380} \\
 Video Pinball     &      0.0 & 17667.9 &           \textbf{811610.0} & 685911.0 & 42684\\
 Wizard Of Wor     &    563.5 &  4756.5 &             6804.7 &   \textbf{7655.7} & 3393 \\
 Zaxxon            &     32.5 &  9173.3 &            11491.7 &  \textbf{12947.6} & 4977 \\
\hline
\end{tabular}
\caption{Maximal evaluation Scores achieved by agents}
\label{table: results final}
\end{table}
\end{center}
}

\newpage

We now compare our method against the results in \cite{stadie2015incentivizing}.
In this paper they introduce a new measure of performance called AUC-100, which is something similar to normalized cumulative rewards up to 20 million frames.
Table \ref{table: auc} displays the results for our reference DQN and bootstrapped DQN as Boot-DQN.
We reproduce their reference results for DQN as DQN* and their best performing algorithm, Dynamic AE.
We also present bootstrapped DQN without head rescaling as Boot-DQN+.

\begin{center}
\begin{table}[!h]
\centering
\begin{tabular}{lccccc}
\hline
                   &   DQN* &   Dynamic AE &   DQN &  Boot-DQN &   Boot-DQN+ \\
\hline
 Alien             &         0.15 &     0.20 &  0.23 &    0.23 &         \textbf{0.33} \\
 Asteroids         &         0.26 &     0.41 &  0.29 &    0.29 &         \textbf{0.55} \\
 Bank Heist        &         0.07 &     0.15 &  0.06 &    0.09 &         \textbf{0.77} \\
 Beam Rider        &         0.11 &     0.09 &  0.24 &    0.46 &         \textbf{0.79} \\
 Bowling           &         0.96 &     \textbf{1.49} &  0.24 &    0.56 &         0.54 \\
 Breakout          &         0.19 &     0.20 &  0.06 &    0.16 &         \textbf{0.52} \\
 Enduro            &         0.52 &     0.49 &  1.68 &    \textbf{1.85} &         1.72 \\
 Freeway           &         0.21 &     0.21 &  0.58 &    0.68 &         \textbf{0.81} \\
 Frostbite         &         0.57 &     0.97 &  0.99 &    \textbf{1.12} &         0.98 \\
 Montezuma Revenge &         0.00 &     0.00 &  0.00 &    0.00 &         0.00 \\
 Pong              &         0.52 &     0.56 & -0.13 &   0.02 &         \textbf{0.60} \\
 Qbert             &         0.15 &     0.10 &  0.13 &    0.16 &         \textbf{0.24} \\
 Seaquest          &         0.16 &     0.17 &  0.18 &    0.23 &         \textbf{0.44} \\
 Space Invaders    &         0.20 &     0.18 &  0.25 &    0.30 &         \textbf{0.38} \\
\hline
 \textbf{Average}  &         0.29 &     0.37 &  0.35 &    0.41 &         \textbf{0.62} \\
\hline
\end{tabular}
\caption{AUC-100 for different agents compared to \cite{stadie2015incentivizing}}
\label{table: auc}
\end{table}
\end{center}

We see that, on average, both bootstrapped DQN implementations outperform Dynamic AE, the best algorithm from previous work.
The only game in which Dynamic AE produces best results is Bowling, but this difference in Bowling is dominated by the implementation of DQN* vs DQN.
Bootstrapped DQN still gives over 100\% improvement over its relevant DQN baseline.
Overall it is clear that Boot-DQN+ (bootstrapped DQN without rescaling) performs best in terms of AUC-100 metric.
Averaged across the 14 games it is over 50\% better than the next best competitor, which is bootstrapped DQN with gradient normalization.

However, in terms of peak performance over 200m frames Boot-DQN generally reached higher scores.
Boot-DQN+ sometimes plateaud early as in Figure \ref{fig: norm plateau}.
This highlights an important distinction between evaluation based on best learned policy versus cumulative rewards, as we discuss in Appendix \ref{app: grad norm}.
Bootstrapped DQN displays the biggest improvements over DQN when doing well during learning is important.

\end{document}